\documentclass[runningheads]{llncs}
\usepackage[T1]{fontenc}
\usepackage{amsmath,amssymb}

\usepackage{color}
\usepackage{tikz}
\usepackage{epsfig}
\usepackage{lipsum}
\usepackage{float}
\usepackage{stfloats}
\usepackage{multicol}
\usepackage{multirow}
\usepackage{bm}
\usepackage{bbm}
\usepackage{etoolbox}
\usepackage{array}
\usepackage{tabulary}
\usepackage{textcomp}
\usepackage{gensymb}
\usepackage{paralist}
\usepackage{booktabs}
\usepackage{adjustbox}
\usepackage{icomma}

\usepackage[pagebackref=true,breaklinks=true,colorlinks,bookmarks=true,bookmarksopen=true]{hyperref}

\definecolor{redarrow}{HTML}{FF6666}
\definecolor{greenarrow}{HTML}{4D9900}
\newcommand{\green}[1]{\textcolor[RGB]{96,177,87}{#1}}

\newcommand{\fn}[1]{{#1}}
\newcommand{\gbf}[1]{\green{\bf{\fn{\scriptsize (#1)}}}}

\usepackage{tikz}
\newcommand*\circled[1]{\tikz[baseline=(char.base)]{
\node[shape=circle,fill=gray,inner sep=0.5pt] (char) {\textcolor{white}{\footnotesize \textbf{#1}}};}}

\definecolor{cgreen}{RGB}{26, 110, 53}
\definecolor{cgrass}{RGB}{123, 252, 3}
\definecolor{cbrown}{RGB}{161, 100, 56}
\definecolor{cyellow}{RGB}{237, 187, 36}
\definecolor{cpurple}{RGB}{177, 87, 250}
\definecolor{cpurblue}{RGB}{194, 207, 242}
\definecolor{cgrey}{RGB}{157, 163, 163}
\definecolor{corange}{RGB}{245, 130, 69}
\definecolor{cblue}{RGB}{66, 120, 245}
\definecolor{csky}{RGB}{148, 250, 255}
\definecolor{ccyan}{RGB}{8, 189, 171}
\definecolor{crose}{RGB}{235, 101, 157}
\definecolor{cpink}{RGB}{255, 212, 212}
\definecolor{cred}{RGB}{219, 15, 15}
\definecolor{cdark}{RGB}{0, 0, 0}
\definecolor{cpink}{RGB}{255, 212, 212}
\usepackage[pdftex]{pict2e}
\newcommand{\thickline}{$\mathrel{\vcenter{\hbox{\rule{15pt}{2pt}}}}$ }

\newcommand{\cstick}[1]{ {\color{#1} \thickline} }

\usepackage{xspace}
\usepackage{graphicx}

\usepackage{algorithm}
\usepackage{listings}
\usepackage{icomma}

\makeatletter
\DeclareRobustCommand\onedot{\futurelet\@let@token\@onedot}
\def\@onedot{\ifx\@let@token.\else.\null\fi\xspace}
\def\eg{\emph{e.g}\onedot} 
\def\ie{\emph{i.e}\onedot} 
 
\def\etc{\emph{etc}\onedot} 
\def\wrt{w.r.t\onedot} 

\makeatother

\makeatletter
\renewcommand*{\@fnsymbol}[1]{\ensuremath{\ifcase#1\or *\or \dagger\or \ddagger\or
    \mathsection\or \mathparagraph\or \|\or **\or \dagger\dagger
    \or \ddagger\ddagger \else\@ctrerr\fi}}
\makeatother

\begin{document}
\title{MatchFormer: Interleaving Attention in Transformers for Feature Matching}

\author{Qing Wang\thanks{Equal contribution} \and Jiaming Zhang\inst{*} \and Kailun Yang\thanks{Correspondence: kailun.yang@kit.edu} \and\\Kunyu Peng \and Rainer Stiefelhagen}

\authorrunning{Q. Wang \emph{et al.}}
\institute{Karlsruhe Institute of Technology, Germany\\
\url{https://github.com/jamycheung/MatchFormer}
}
\maketitle
\begin{abstract}
\sloppy 
Local feature matching is a computationally intensive task at the subpixel level. While \emph{detector-based} methods coupled with feature descriptors struggle in low-texture scenes, CNN-based methods with a sequential \emph{extract-to-match} pipeline, fail to make use of the matching capacity of the encoder and tend to overburden the decoder for matching. In contrast, we propose a novel hierarchical \emph{extract-and-match} transformer, termed as \emph{MatchFormer}.  Inside each stage of the hierarchical encoder, we interleave self-attention for feature extraction and cross-attention for feature matching, yielding a human-intuitive \emph{extract-and-match} scheme. Such a match-aware encoder releases the overloaded decoder and makes the model highly efficient. Further, combining self- and cross-attention on multi-scale features in a hierarchical architecture improves matching robustness, particularly in low-texture indoor scenes or with less outdoor training data. Thanks to such a strategy, MatchFormer is a multi-win solution in efficiency, robustness, and precision. Compared to the previous best method in indoor pose estimation, our lite MatchFormer has only $45\%$ GFLOPs, yet achieves a $+1.3\%$ precision gain and a $41\%$ running speed boost. The large MatchFormer reaches state-of-the-art on four different benchmarks, including indoor pose estimation (ScanNet), outdoor pose estimation (MegaDepth), homography estimation and image matching (HPatch), and visual localization (InLoc).

\end{abstract}
\section{Introduction}

\begin{figure}[t]
    \begin{center}
      \includegraphics[width=1\linewidth, keepaspectratio]{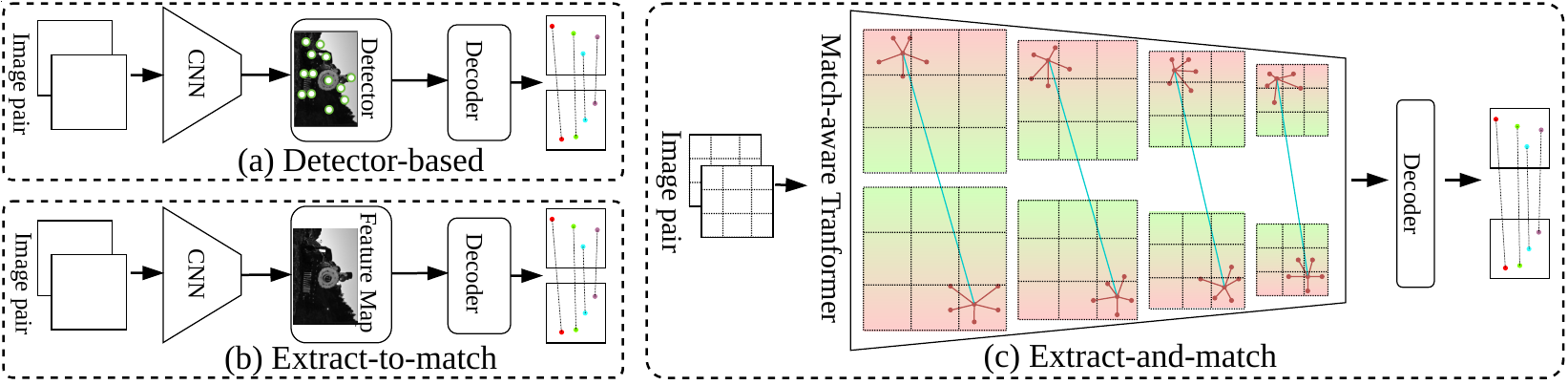}
    \end{center}
    \vskip -4ex
    \caption{\small \textbf{ 
    Feature matching pipelines.} While (a)~\emph{detector-based} methods coupled with feature descriptors, (b)~\emph{extract-to-match} methods fail to make use of the matching capacity of the encoder. 
    Self- and cross-attention are interleaved inside each stage of the match-aware transformer to perform a novel (c)~\emph{extract-and-match} pipeline.}
    \label{fig:pipeline}
\vskip -2ex
\end{figure}

Matching two or more views of a scene is the core of many basic computer vision tasks, \eg, Structure-from-Motion (SfM)~\cite{lindenberger2021pixel_perfect,ma2022virtual_correspondence}, Simultaneous Localization and Mapping (SLAM)~\cite{chen2021panoramic,engel2017direct}, relative pose estimation~\cite{li2015dual}, and visual localization~\cite{sarlin2019coarse,taira2018inloc,yoon2021line}, \etc. For vision-based matching, classical \emph{detector-based} methods (see Fig.~\ref{fig:pipeline}(a)), coupled with hand-crafted local features~\cite{dusmanu2019d2,rublee2011orb}, are computationally intensive due to the high dimensionality of local features~\cite{sarlin2019coarse,zhou2020integrating}.
Recent works~\cite{luo2019contextdesc,revaud2019r2d2,wang2020learning} based on deep learning focus on learning detectors and local descriptors using Convolutional Neural Networks (CNNs). Some partial transformer-based methods~\cite{sun2021loftr,jiang2021cotr} only design an attention-based decoder and remain the \emph{extract-to-match} pipeline (see Fig.~\ref{fig:pipeline}(b)).
For instance, while COTR~\cite{jiang2021cotr} feeds CNN-extracted features into a transformer-based decoder, SuperGlue~\cite{sarlin2020superglue} and LoFTR~\cite{sun2021loftr} only apply attention modules atop the decoder. Overburdening the decoder, yet neglecting the matching capacity of the encoder, makes the whole model computationally inefficient. 

\begin{figure}[t]
    \begin{center}
       \includegraphics[width=1\linewidth, keepaspectratio]{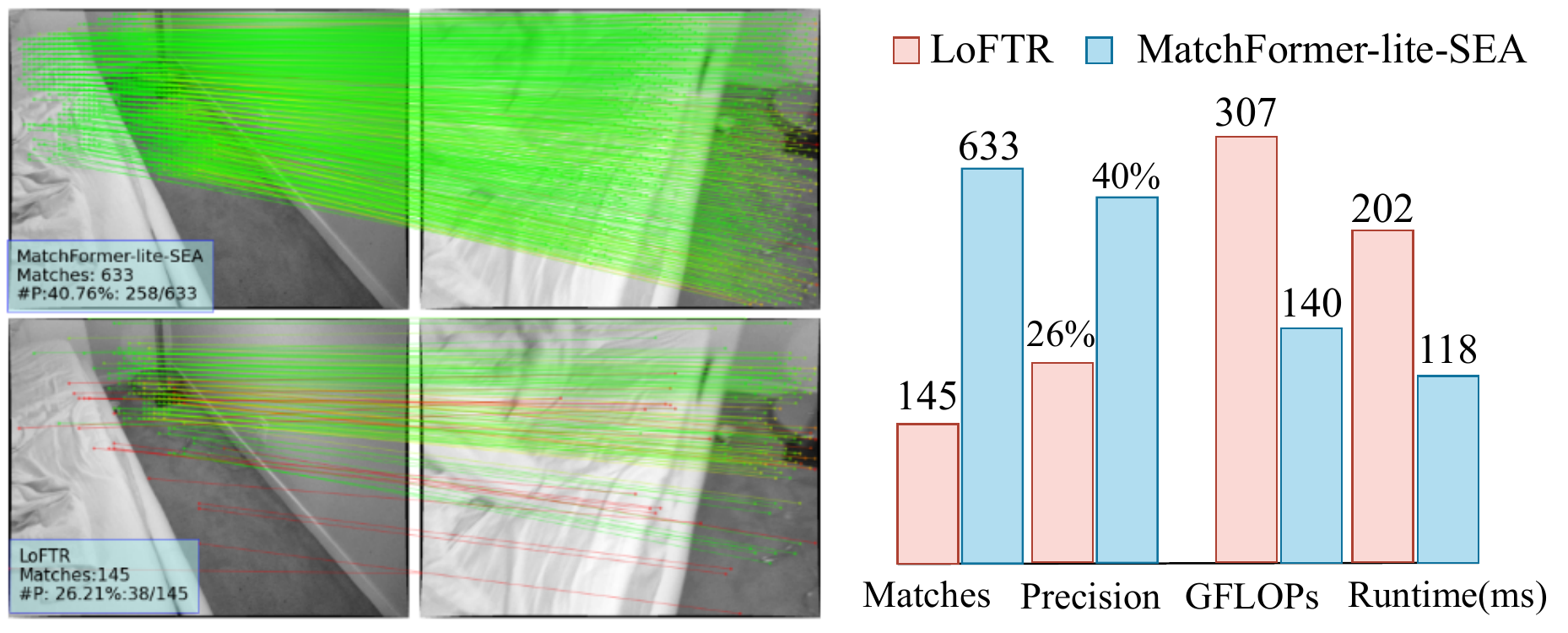}
    \end{center}
    \vskip -4ex
    \caption{\small \textbf{ Comparison between MatchFormer and LoFTR.} With $45\%$ GFLOPs of LoFTR, our {efficient} MatchFormer boosts the running speed by $41\%$, while delivering more {robust} matches and a higher matching {precision} on such a low-texture indoor scenario. Green color in the figure refers to correct matches and red color to mismatches.} 
    \label{fig:main}
\vskip -4ex
\end{figure}

\emph{Rethinking local feature matching, in reality, one can perform feature extraction and matching simultaneously by using a pure transformer.} We propose an \emph{extract-and-match} pipeline shown in Fig.~\ref{fig:pipeline}(c).
Compared to the \emph{detector-based} methods and the \emph{extract-to-match} pipeline, our new scheme is more in line with human intuition, which learns more respective features of image pairs while paying attention to their similarities~\cite{zhong2020subspace}. To this end, a novel transformer termed \emph{MatchFormer} is proposed, which helps to achieve multi-wins in precision, efficiency, and robustness of feature matching. For example, compared to LoFTR~\cite{sun2021loftr} in Fig.~\ref{fig:main}, MatchFormer with lower GFLOPs is more robust in low-textured scenes and achieves higher matching number, speed, and accuracy. 

More specifically, for improving computational efficiency and the robustness in matching low-texture scenes, we put forward \emph{interleaving} self- and cross-attention in MatchFormer to build a matching-aware encoder. In this way, the local features of the image itself and the similarities of its paired images can be learned simultaneously, so called \emph{extract-and-match}, which relieves the overweight decoder and makes the whole model efficient. The cross-attention arranged in earlier stages of the encoder robustifies feature matching, particularly, in low-texture indoor scenarios or with less training samples outdoors, which makes MatchFormer more suitable for real-world applications where large-scale data collection and annotation are infeasible.
To extract continuous patch information and embed location information, a novel \emph{positional patch embedding~(PosPE)} method is designed in the matching-aware encoder, which can enhance the detection of low-level features. Additionally, the lite and large versions \textit{w.r.t.} feature resolutions, each with two efficient attention modules~\cite{shen2021efficient,wang2021pvt}, are fully investigated to overcome the massive calculations in transformers when dealing with fine features. Furthermore, MatchFormer, with a hierarchical transformer, conducts multi-level feature extraction in the encoder and multi-scale feature fusion in the decoder, which contribute to the robustness of matching. Finally, for the precision, extensive experiments prove that MatchFormer achieves state-of-the-art performances of indoor location estimation on ScanNet~\cite{dai2017scannet}, outdoor location estimation on MegaDepth~\cite{Li_2018_CVPR}, image matching and homography estimation on HPatches~\cite{balntas2017hpatches}, and visual localization on InLoc~\cite{taira2018inloc}. 

In summary, the contributions of this paper include:
\begin{compactitem}
    \item We rethink local feature matching and propose a new \emph{extract-and-match} pipeline, which enables synchronization of feature extraction and feature matching. The optimal combination path is delivered when \emph{interleaving} self- and cross-attention modules within each stage of the hierarchical structure to enhance multi-scale features. 
    \item We propose a novel vision transformer, \ie, \emph{MatchFormer}, equipped with a robust hierarchical transformer encoder and a lightweight decoder. Including lite and large versions and two attention modules, four variants of MatchFormer are investigated.
    \item We introduce a simple and effective positional patch embedding method, \ie, \emph{PosPE}, which can extract continuous patch information and embed location information, as well as enhances the detection of low level features.
    \item MatchFormer achieves state-of-the-art scores on matching low-texture indoor images and is superior to previous \emph{detector-based} and \emph{extract-to-match} methods in pose estimation, homography estimation, and visual localization.
\end{compactitem}

\section{Related Work}
\noindent\textbf{Local Feature Matching.}
\emph{Detector-based} methods~\cite{dusmanu2019d2,cheng2018visual_localization,fang2020cfvl,luo2020aslfeat} usually include five steps: detecting interest points, calculating visual descriptors, searching for nearest neighbor matches, rejecting incorrect matches, and estimating geometric transformations.
In \emph{extract-to-match} methods~\cite{dusmanu2019d2,revaud2019r2d2,sun2021loftr,sarlin2020superglue,li2020dual,tang2022quadtree,shi2022clustergnn,revaud2022pump} designed for feature matching, CNNs are normally adopted to learn dense and discriminative features. CAPS~\cite{wang2020learning} fuses multi-resolution features extracted by CNNs and obtains the descriptor of each pixel through interpolation.
DSM~\cite{tang2021learning} strengthens detection and refines the descriptors by merging various frames and multiple scales extracted by CNNs.
DRC-Net~\cite{li2020dual} obtains CNN feature maps of two different resolutions, generates two 4D matching tensors, and fuses them to achieve high-confidence feature matching.
D2Net~\cite{dusmanu2019d2} obtains valid key points by detecting the local maximum of CNN features.
R2D2~\cite{revaud2019r2d2} adapts dilated convolutions~\cite{chen2017deeplab,yu2015multi_dilated} to maintain image resolution and predict each key points and descriptors. 
COTR~\cite{jiang2021cotr}, LoFTR~\cite{sun2021loftr}, and QuadTree~\cite{tang2022quadtree} follow sequential \emph{extract-to-match} processing. 
In this work, we consider that feature extraction and similarity learning through a transformer synchronously, can provide matching-aware features in each stage of the hierarchical structure. 

\noindent\textbf{Vision Transformer.}
Transformer~\cite{dosovitskiy2020image} excels at capturing long-distance dependency~\cite{vaswani2017attention}, making it outstanding in vision tasks such as classification~\cite{liu2021swin,touvron2021deit,yuan2021t2t}, detection~\cite{carion2020detr,wang2021pvt,zhu2020deformable_detr}, semantic segmentation~\cite{zheng2021setr,xie2021segformer,zhang2021trans4trans}, image enhancement~\cite{zhang2021star}, and image synthesis~\cite{esser2021taming}.
For local-feature matching, only attention blocks of transformers have been used in recent works.
For example, SuperGlue~\cite{sarlin2020superglue} and LoFTR~\cite{sun2021loftr} applied self- and cross-attention to process the features which were extracted from CNNs.
Yet, attention can actually function as the backbone module for feature extraction instead of only being used in the decoder for CNNs. This has been verified in ViT~\cite{dosovitskiy2020image}, but mainly for classification and segmentation tasks~\cite{zheng2021setr,wang2021pvt}.
It remains unclear whether it is transferable to the image feature matching.
When a pure transformer framework is used to process local feature matching, the computation complexity will be exceedingly large. Besides, transformers often lack and miss local feature information~\cite{yuan2021t2t}.
In this paper, we put forward a fully transformer image matching framework.
In our model, we design positional patch embedding to enhance the feature extraction and introduce interleaving attention to achieve efficient and robust feature matching.

\section{Methodology}

\begin{figure*}[t]
\begin{center}
    \includegraphics[width= 1.0\linewidth, keepaspectratio]{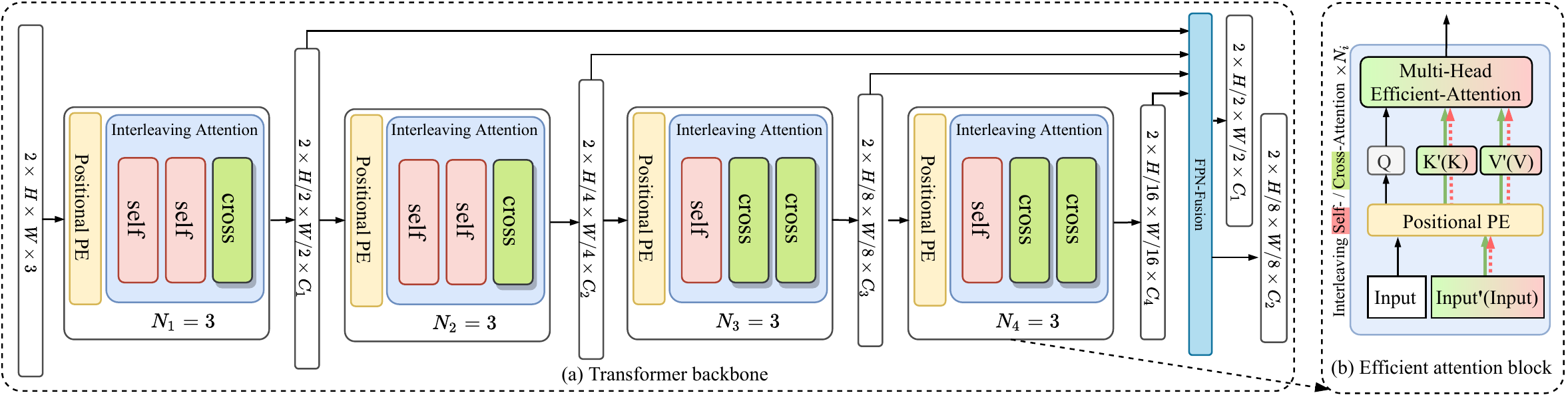}
\end{center}
\vskip -4ex
   \caption{\small \textbf{MatchFormer architecture}: (a) The transformer backbone generates high-resolution coarse features and low-resolution fine features; 
   In (b), each attention block has interleaving-arranged self-attention (\textit{w.r.t.} $\boldsymbol{Q}$, $\boldsymbol{K}$, $\boldsymbol{V}$ and \textcolor{redarrow}{red} arrows) within the $input$, and cross-attention (\textit{w.r.t.} $\boldsymbol{Q}$, $\boldsymbol{K'}$, $\boldsymbol{V'}$ and alternative \textcolor{greenarrow}{green} arrows) cross images ($input$ and $input'$). Multi-head efficient-attention reduces the computation; Positional Patch Embedding~(PE) completes the patch embedding and the position encoding. 
   }
\label{fig:match}
\vskip -4ex
\end{figure*}

\subsection{MatchFormer}
As illustrated in Fig.~\ref{fig:match}, MatchFormer employs a hierarchical transformer, which comprises four stages to generate high-resolution coarse and low-resolution fine features for local feature matching. In four stages, the self- and cross-attention are arranged in an \emph{interleaving} strategy. Each stage consists of two components: one \emph{positional patch embedding~(PosPE)} module, and a set of efficient attention modules. Then, the multi-scale features are fused by an FPN-like decoder. Finally, the coarse and fine features are passed to perform the coarse-to-fine matching, as introduced in LoFTR~\cite{sun2021loftr}.

\noindent\textbf{Extract-and-Match Pipeline.} 
Unlike the \emph{extract-to-match} LoFTR using attention on a single-scale feature map and only after feature extraction, we combine self- and cross-attention inside the transformer-based encoder and apply on multiple feature scales (see Fig.~\ref{fig:pipeline}). The combination of two types of attention modules enables the model to extract non-local features via self-attention and explore their similarities via cross-attention simultaneously, so called the \emph{extract-and-match} scheme. As a new matching scheme, however, the difficulty lies in finding an effective and optimal combination strategy while maintaining the efficiency and robustness of the entire model. Thanks to the hierarchy nature of Transformers~\cite{dosovitskiy2020image,wang2021pvt}, we obtain two insights: (1) As the feature map at the shallow stage emphasizes textural information, \emph{relatively} more self-attention are applied to extract the feature itself on the early stages. (2) As the feature map at the deep stage is biased toward semantic information, \emph{relatively} more cross-attention are developed to explore the feature similarity on the later stages. These two observations lead us to design a novel \emph{interleaving} strategy for joining self- and cross-attention. 

\noindent\textbf{Interleaving Self-/Cross-Attention.} As shown in Fig.~\ref{fig:match}(a), the combination of self- and cross-attention modules are set at each stage in an \emph{interleaving} strategy. Each block in Fig.~\ref{fig:match}(b) contains $N$ attention modules, where each attention module is represented as self-attention or alternative cross-attention according to the input image pair. For self-attention, $\boldsymbol{Q}$ and ($\boldsymbol{K}$, $\boldsymbol{V}$) come from the same \emph{input}, so the self-attention is responsible for feature extraction of the image itself. For cross-attention, ($\boldsymbol{K'}$, $\boldsymbol{V'}$) are from another $input'$ of the image pair. Thus, the cross-attention learns the similarity of the image pair, resulting a match-aware transformer-based encoder. 
Within an attention block, self-attended features are extracted, while the similarity of the feature pair is located by the cross-attention. The strategy is more human-intuitive, which learns more respective features of image pairs while paying attention to their similarities. 

\begin{figure}[t]
\begin{center}
   \includegraphics[width=0.99\linewidth, keepaspectratio]{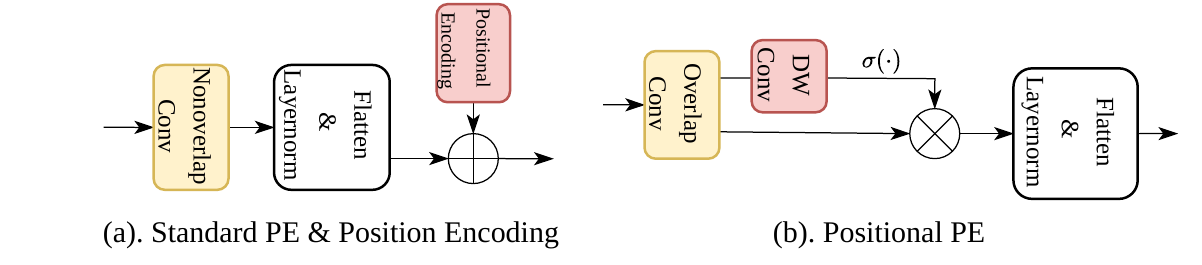}
\end{center}
    \vskip -6ex
       \caption{ \small \textbf{Comparison between different patch embedding modules.}
       } 
    \label{fig:pospe}
\vskip -4ex
\end{figure}

\noindent\textbf{Positional Patch Embedding~(PosPE).} 
Typical transformers~\cite{dosovitskiy2020image}, split the image $(H{\times}W{\times}3)$ into patches with size of $P{\times}P$ and then flatten these patches into sequence with a size of $N{\times}C$, where $N{=}HW / P^2$.  
The process is difficult to gather location information around patches.
As a result, low-level feature information cannot be acquired directly through the standard process~\cite{yuan2021t2t}, which severely restricts the local feature matching. In the case of standard Patch Embeding~(PE) in Fig.~\ref{fig:pospe}(a), the independent patch ignores the information around it and requires additional position encoding at the end. Therefore, we propose a simple but effective positional patch embedding~(PosPE) method for capturing low feature information with few parameters, as shown in Fig.~\ref{fig:pospe}(b). It has a $7{\times}7$ convolution layer (with padding $3$ and stride $2$) in the first stage, and $3{\times}3$ convolution layers (all with padding $1$ and stride $2$) in later stages. 
A depth-wise $3{\times}3$ convolution
is added to further enhance local features and encode positional information by its padding operation. The pixel-wise weights are then scaled by a sigmoid function $\sigma(\cdot)$ after the first step of convolution. 
Besides, our PosPE includes a first overlapping convolution that captures the continuous patch area information. 
PosPE augments the location information of patches and extracts denser features, which facilitates accurate feature matching.

\noindent\textbf{Preliminaries on Efficient-Attention.} After Patch Embedding, the query $\boldsymbol{Q}$, key $\boldsymbol{K}$, and value $\boldsymbol{V}$ are obtained, with the same $N{\times}C$ dimension according to the input resolution $N{=}H{\times}W$.
The computation of the traditional attention is formulated as: $softmax((\boldsymbol{Q}\boldsymbol{K}^T)/{\sqrt{d}})\boldsymbol{V}$,
where $\sqrt{d}$ is the scaling factor.
However, the product of $\boldsymbol{Q}\boldsymbol{K}^T$ introduces a $O(N^2)$ complexity, which is prohibitive in large image resolutions and makes the model inefficient.
To remedy this problem, we apply two kinds of efficient attention, \ie, \textit{Spatial Efficient Attention~(SEA)} as in ~\cite{wang2021pvt,xie2021segformer} or \textit{Linear Attention (LA)} as in~\cite{shen2021efficient}. Then, $O(N^2)$ is reduced to $O({N^2}/{R})$ or $O(N)$. Hence, larger input feature maps can be well handled and processed while using a pure transformer-based encoder in the feature matching task. 

\noindent\textbf{Multi-scale Feature Fusion.}
Apart from the interleaving combination, there are four different stages in our hierarchical transformer encoder, in which the feature resolution shrinks progressively. 
Different from previous works~\cite{li2020dual,jiang2021cotr} considering only the single-scale feature, MatchFormer fuses multi-scale features to generate dense and match-aware features for feature matching. 
As shown in Fig.~\ref{fig:match}(a), we flexibly adopt an FPN-like decoder in our architecture, 
because it can bring two benefits: (1) generating more robust coarse- and fine features for promoting the final matching; (2) creating a lightweight decoder without making the whole model computationally complex. 
\vskip-3ex

\begin{table}[t]
    \centering
    \caption{\textbf{MatchFormer-lite and -large with Linear Attention~(LA) and Spatial Efficient Attention~(SEA).} $C$: the channel number of feature $\boldsymbol{F}$; $K$, $S$ and $P$: the patch size, stride, and padding size of PosPE; 
    $E$: the expansion ratio of MLP in an attention block;
    $A$: the head number of attention; $R$: the down-scale ratio of SEA.
    }
    \vskip -2ex
    \resizebox{0.98\columnwidth}{!}{
\renewcommand{\arraystretch}{0.9}
\setlength\tabcolsep{8.0pt}
\begin{tabular}{c|c|c|c|c|c}
\toprule
\textbf{Stage} & \multicolumn{2}{c|}{\textbf{MatchFormer-lite}} & \multicolumn{2}{c|}{\textbf{MatchFormer-large}} & \textbf{$N_{i}$}\\
\hline \hline

\multirow{4}{*}{$\boldsymbol{F}_{1}$} &\multirow{2}{*}{$H/4{\times}W/4$} &\multirow{2}{*}{$K$=7, $S$=4, $P$=3, $E$=4} &\multirow{2}{*}{$H/2{\times}W/2$}  &\multirow{2}{*}{$K$=7, $S$=2, $P$=3, $E$=4}& \\ 
& & & &\\ \cline{3-3} \cline{5-6}
&\multirow{2}{*}{$C_1$=128} &\multirow{2}{*}{\textbf{LA:} $A$=8 ; \textbf{SEA:} $A$=1, $R$=4} &\multirow{2}{*}{$C_1$=128} &\multirow{2}{*}{\textbf{LA:} $A$=8 ; \textbf{SEA:} $A$=1, $R$=4} &\multirow{2}{*}{${\times}3$} \\ 
& & & &\\

\hline

\multirow{4}{*}{$\boldsymbol{F}_{2}$} &\multirow{2}{*}{$H/8{\times}W/8$} &\multirow{2}{*}{$K$=3, $S$=2, $P$=1, $E$=4} &\multirow{2}{*}{$H/4 {\times}W/4$}  &\multirow{2}{*}{$K$=3, $S$=2, $P$=1, $E$=4}& \\
& & & &\\ \cline{3-3} \cline{5-6}
&\multirow{2}{*}{$C_2$=192} &\multirow{2}{*}{\textbf{LA:} $A$=8 ; \textbf{SEA:} $A$=2, $R$=2} &\multirow{2}{*}{$C_2$=192} &\multirow{2}{*}{\textbf{LA:} $A$=8 ; \textbf{SEA:} $A$=2, $R$=2} &\multirow{2}{*}{${\times}3$} \\ 
& & & & & \\ 

\hline

\multirow{4}{*}{$\boldsymbol{F}_{3}$} &\multirow{2}{*}{$H/16{\times}W/16$} &\multirow{2}{*}{$K$=3, $S$=2, $P$=1, $E$=4} &\multirow{2}{*}{$H/8 {\times}W/8$}  &\multirow{2}{*}{$K$=3, $S$=2, $P$=1, $E$=4}& \\ 
& & &&\\ \cline{3-3} \cline{5-6}
&\multirow{2}{*}{$C_3$=256} &\multirow{2}{*}{\textbf{LA:} $A$=8 ; \textbf{SEA:} $A$=4, $R$=2} &\multirow{2}{*}{$C_3$=256} &\multirow{2}{*}{\textbf{LA:} $A$=8 ; \textbf{SEA:} $A$=4, $R$=2} &\multirow{2}{*}{${\times}3$} \\ 
& & & & & \\ 

\hline

\multirow{4}{*}{$\boldsymbol{F}_{4}$} &\multirow{2}{*}{$H/32{\times}W/32$} &\multirow{2}{*}{$K$=3, $S$=2, $P$=1, $E$=4} &\multirow{2}{*}{$H/16 {\times}W/16$}  &\multirow{2}{*}{$K$=3, $S$=2, $P$=1, $E$=4}& \\ 
& & &&\\ \cline{3-3} \cline{5-6}
&\multirow{2}{*}{$C_4$=512} &\multirow{2}{*}{\textbf{LA:} $A$=8 ; \textbf{SEA:} $A$=8, $R$=1} &\multirow{2}{*}{$C_4$=512} &\multirow{2}{*}{\textbf{LA:} $A$=8 ; \textbf{SEA:} $A$=8, $R$=1} &\multirow{2}{*}{${\times}3$} \\ 
& & & & & \\

\hline

\multirow{2}{*}{\textbf{Output}}  & \multicolumn{2}{c|}{Coarse: $H/4 {\times}W/4, 128$} & \multicolumn{2}{c|}{Coarse: $H/2 {\times}W/2, 128$} & \\
& \multicolumn{2}{c|}{Fine: $H/8 {\times}W/8, 192$} & \multicolumn{2}{c|}{Fine: $H/8 {\times}W/8, 256$} & \\
\bottomrule
\end{tabular}
}
    \label{tab:strct}
    \vskip -4ex
\end{table}

\subsection{Model Settings}
\noindent\textbf{MatchFormer Variants.}  
MatchFormer is available with its \emph{lite} and \emph{large} versions, as presented in Table~\ref{tab:strct}. For the MatchFormer-lite models, we pick a lower resolution setting, which greatly increases the matching efficiency and ensures a certain matching accuracy. Therefore, we set MatchFormer-lite $4$-stage features in the respective resolution of $\frac{1}{r_i}{\in}\{\frac{1}{4},\frac{1}{8},\frac{1}{16},\frac{1}{32}\}$ of the input. To promote context learning for matching, feature embeddings with higher channel numbers are beneficial, which are set as $C_i{\in}\{128,192,256,512\}$ for four stages. In the MatchFormer-large models, higher resolution feature maps facilitate accurate dense matching. Hence, the $\frac{1}{r_i}$ and $C_i$ are set as $\{\frac{1}{2},\frac{1}{4},\frac{1}{8},\frac{1}{16}\}$ and $\{128,192,256,512\}$ for the large MatchFormer. 

\noindent\textbf{Attention Module Variants.}  
To fully explore the proposed \emph{extract-and-match} scheme, each of the two MatchFormer variants has two attention variants. Here, we mainly investigate Linear Attention (LA) and Spatial Efficient Attention (SEA). Thus, there are four versions of MatchFormer as presented in Table~\ref{tab:strct}.  
We found that they have different capabilities for recognizing features, making them suitable for various tasks. In the local feature matching, the density of features is different indoors and outdoors. We study the two kinds of attention in indoor~(in Sec.~\ref{sec:indoor_pos}) and outdoor~(in Sec.~\ref{sec:outdoor_pos}) pose estimation, respectively.

\section{Experiments}
\subsection{Implementation and Datasets}
\vskip-1ex

\noindent\textbf{ScanNet.}
We use ScanNet~\cite{dai2017scannet} to train our indoor models. ScanNet is an indoor RGB-D video dataset with $2.5$ million views in $1,513$ scans with ground-truth poses and depth maps. The lack of textures, the ubiquitous self-similarity, and the considerable changes in viewpoint make ScanNet a challenging dataset for indoor image matching. Following~\cite{sarlin2020superglue}, we select $230$ million image pairs with the size of $640{\times}480$ as the training set and $1,500$ pairs as the testing set.

\noindent\textbf{MegaDepth.}
Following~\cite{dusmanu2019d2}, we use MegaDepth~\cite{Li_2018_CVPR} to train our outdoor models, which has $1$ million internet images of $196$ scenarios, and their sparse 3D point clouds are created by COLMAP~\cite{schonberger2016structure}. 
We use $38,300$ image pairs from $368$ scenarios for training, and the same $1,500$ testing pairs from~\cite{sun2021loftr} for evaluation.

\noindent\textbf{Implementation Settings.}
On the indoor dataset ScanNet, MatchFormer is trained using Adam~\cite{kingma2014adam} with initial learning rate and batch size, setting for the lite version at $3 {\times} 10^{-3}$ and $4$, and for the large version at $3 {\times} 10^{-4}$ and $2$.
In the case of the outdoor dataset MegaDepth, MatchFormer is trained using Adam with initial learning rate and batch size, setting for the lite version at $3 {\times} 10^{-3}$ and $2$, and for the large version at $3 {\times} 10^{-4}$ and $1$.
To compare LoFTR and MatchFormer at different data scales on outdoor pose estimation task, both use 8 A100 GPUs, otherwise use 64 A100 GPUs following LoFTR~\cite{sun2021loftr}. 
We perform Image Matching, Homography Estimation, and InLoc Visual Localization experiments using the model trained with MatchFormer-large-LA on MegaDepth.

\begin{table}[t]
    \centering
    \vskip -2ex
    \caption{\textbf{Indoor pose estimation on ScanNet.} 
    The AUC of three different thresholds and the average matching precision (P) are evaluated.
    }
    \resizebox{0.9\columnwidth}{!}{
\setlength\tabcolsep{14.0pt}
\renewcommand{\arraystretch}{0.85}
\begin{tabular}{lcccc}
    \toprule
    \multicolumn{1}{c}{\multirow{2}{*}{Method}}& \multicolumn{3}{c}{Pose estimation AUC (\%)} & \multicolumn{1}{c}{\multirow{2}{*}{P}} \\
    \cline{2-4} & \multirow{1}{*}{@5\degree} & \multirow{1}{*}{@10\degree} & \multirow{1}{*}{@20\degree} &          \\
    \midrule \midrule
    ORB~\cite{rublee2011orb}+GMS~\cite{bian2017gms}~{\tiny CVPR'17} & 5.21    & 13.65      & 25.36 &   72.0 \\
    D2-Net~\cite{dusmanu2019d2}+NN~{\tiny CVPR'19}& 5.25    & 14.53      & 27.96  &  46.7 \\
    ContextDesc~\cite{luo2019contextdesc}+RT~\cite{lowe2004distinctive}~{\tiny CVPR'19} & 6.64    & 15.01      & 25.75 &   51.2   \\
    SP~\cite{detone2018superpoint}+NN~{\tiny CVPRW'18} &9.43    & 21.53      & 36.40  &  50.4 \\
    SP~\cite{detone2018superpoint}+PointCN~\cite{yi2018learning}~{\tiny CVPR'18} & 11.40     & 25.47      & 41.41  & 71.8  \\
    SP~\cite{detone2018superpoint}+OANet~\cite{zhang2019learning}~{\tiny ICCV'19} & 11.76     & 26.90      & 43.85   & 74.0 \\
    SP~\cite{detone2018superpoint}+SuperGlue~\cite{sarlin2020superglue}~{\tiny CVPR'20} & 16.16     & 33.81      & 51.84   & 84.4 \\
    LoFTR~\cite{sun2021loftr}~{\tiny CVPR'21} &22.06&   40.80  & 57.62 & 87.9 \\
    LoFTR~\cite{sun2021loftr}+QuadTree~\cite{tang2022quadtree}~{\tiny ICLR'22}  & 23.90 & 43.20 & 60.30 & 89.3 \\
    MatchFormer-lite-LA   & 20.42  &39.23   &56.82   & 87.7  \\
    MatchFormer-lite-SEA  & 22.89 &42.68& 60.66& 89.2 \\
    MatchFormer-large-LA   &  24.27   &  43.48 & 60.55  & 89.2  \\
    MatchFormer-large-SEA  & \textbf{24.31}    & \textbf{43.90}  & \textbf{61.41}  & \textbf{89.5}  \\
    \bottomrule
\end{tabular}
}
    \label{tab:inpose}
    \vskip -4ex
\end{table}

\subsection{Indoor Pose Estimation}\label{sec:indoor_pos}
\vskip-1ex
Indoor pose estimation is highly difficult due to wide areas devoid of textures, a high degree of self-similarity, scenes with complicated 3D geometry, and frequent perspective shifts. Faced with all these challenges, MatchFormer with interleaved self- and cross-attention modules still functions well as unfolded in the results.

\noindent\textbf{Metrics.}
Following~\cite{sarlin2020superglue}, we provide the area under the cumulative curve~(AUC) of the pose error at three different thresholds $(5\degree, 10\degree, 20\degree)$. The camera pose is recovered by using RANSAC. We report the matching precision (P), the probability of a true match if its epipolar is smaller than $5{\times}10^{-4}$. 

\noindent\textbf{Quantitative Results.}
As shown in Table~\ref{tab:inpose}, MatchFormer demonstrates exceptional performance on the low-texture indoor pose estimation task. The matching precision (P) of MatchFormer-large-SEA reaches the state-of-the-art level of $89.5\%$. 
Benefiting from the \emph{extract-and-match} strategy, MatchFormer-large-SEA can bring $+5.1\%$ improvement over the \emph{detector-based} SuperGlue, $+1.6\%$ over the \emph{extract-to-match} LoFTR. Pose estimation AUC of MatchFormer is also significantly superior to \emph{detector-based} SuperGlue. 
Compared to LoFTR, MatchFormer provides a more pronounced pose estimation AUC by boosting $(+2.25\%$, $+3.1\%$, $+3.79\%)$ at three thresholds of $(5\degree, 10\degree, 20\degree)$.
The LoFTR model is recently adapted by a complex decoder with QuadTree Attention~\cite{tang2022quadtree}. However, MatchFormer maintains its lead $(+0.41\%, +0.70\%, +1.11\%)$ with the \emph{extract-and-match} strategy. 
Additionally, 
compared to LoFTR, our lightweight MatchFormer-lite-SEA has only $45\%$ GFLOPs, yet achieves a $+1.3\%$ precision gain and a $41\%$ running speed boost. 
More details of the efficiency comparison will be presented in Table~\ref{tab:params}. 
Comparing SEA and LA, we found that the spatial scaling operation in SEA has benefits in handling low-texture features, thus it is more suited for indoor scenes and provides better results.

\begin{figure*}[t]
\begin{center}
    \includegraphics[width= 1.0\linewidth, keepaspectratio]{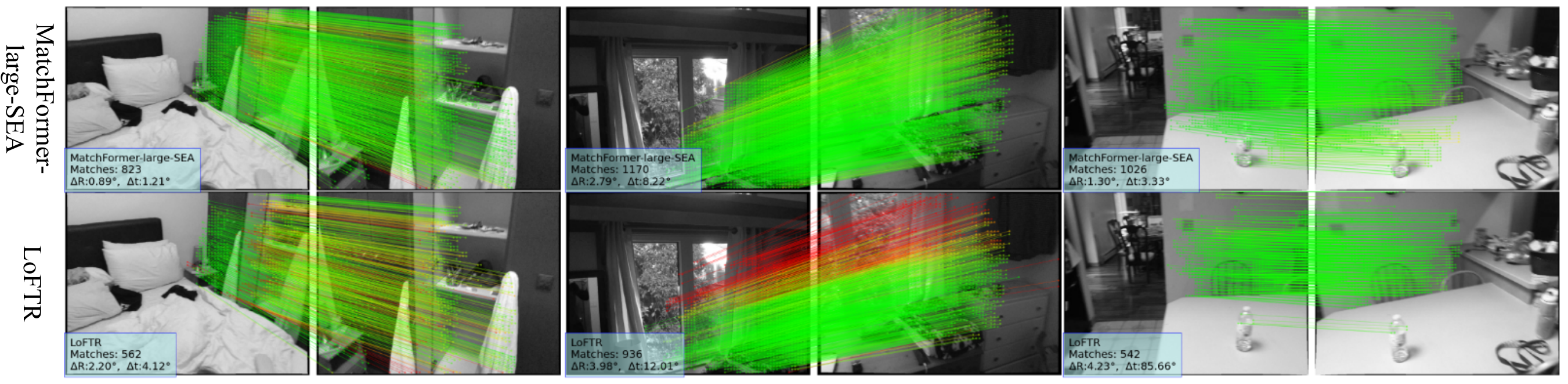}
\end{center}
    \vskip -6ex
   \caption{\small \textbf{Qualitative visualization} of MatchFormer and LoFTR~\cite{sun2021loftr}. 
   MatchFormer achieves higher matching numbers and more correct matches in low-texture scenes.}
\label{fig:matchimage}
\vskip -1ex
\end{figure*}

\noindent\textbf{Qualitative Results.}
The indoor matching results are in Fig.~\ref{fig:matchimage}. In challenging feature-sparse indoor scenes, it can reliably capture global information to assure more matches and high accuracy. Thus, the pose solved by matching prediction has a lower maximum angle error $(\Delta R)$ and translation error $(\Delta t)$.
Due to the hierarchical transformer and interleaving-attention design, the receptive field of MatchFormer exceeds that of CNN-based methods. It confirms that applying cross-attention modules earlier for learning feature similarity robustifies low-texture indoor matching, which is in line with our \emph{extract-and-match} pipeline.

\begin{figure}[t]
    \centering
    \includegraphics[width=1.0\linewidth]{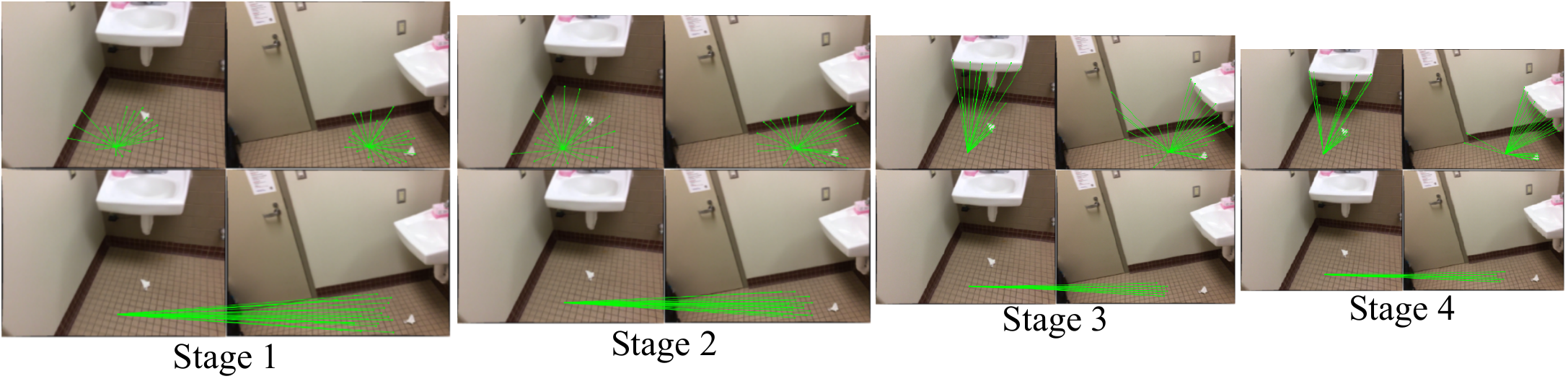}
    \vskip -3ex
    \caption{\small \textbf{Visualization of self- and cross-attention at 4 stages of MatchFormer.} Cross-attention focuses on learning the similarity across paired images and gradually refines the matching range, while self-attention focuses on detecting features of the image itself and enabling long-range dependencies.}
    \label{fig:attention}
    \vskip -4ex
\end{figure}
\noindent\textbf{Self- and Cross-attention Visualization.}
To further investigate the effectiveness of interleaving attention in MatchFormer, the features of self- and cross-attention modules in four stages are shown in Fig.~\ref{fig:attention}.
Self-attention connects obscure points with surrounding points, while cross-attention learns relationship between 
points across images. Specifically, self-attention enables the query point to associate surrounding textural features in the shallow stage, and it enables the query point to connect to semantic features in the deep stage. As the model deepens, cross-attention will narrow the range of query points detected across images, rendering the matching much easier and more fine-grained. Finally, these four stages of features are blended, empowering the model to perform accurate feature matching in low-texture scenes.

\subsection{Outdoor Pose Estimation}\label{sec:outdoor_pos}
\vskip-1ex
Outdoor pose estimation presents unique challenges compared to indoors. In particular, outdoor scenes have greater variations in lighting and occlusion. Still, Matchformer achieves outstanding performance in outdoor scenes.

\noindent\textbf{Metrics.}
We present the same AUC of the pose error 
as in the indoor pose estimation task. 
The matching precision pipolar distance threshold is $1{\times}10^{-4}$.

\begin{table}[t]
    \centering
    \vskip -2ex
    \caption{\textbf{Outdoor pose estimation on MegaDepth.} 
    $\dag$ represents training on different percentages of datasets, which requires 8 GPUs for training. 
    }
    \resizebox{1\columnwidth}{!} {
\small 
\setlength\tabcolsep{4pt}
\renewcommand{\arraystretch}{0.99}
\begin{tabular}{lcllll}
    \toprule
    \multicolumn{1}{c}{\multirow{2}{*}{Method}}   &Data  &                                   \multicolumn{3}{c}{Pose estimation AUC (\%)} & \multicolumn{1}{c}{\multirow{2}{*}{P}} \\
    \cline{3-5}
                                                                     &   percent     & \multirow{1}{*}{@5\degree} & \multirow{1}{*}{@10\degree} & \multirow{1}{*}{@20\degree}           \\
    \midrule \midrule
    SP~\cite{detone2018superpoint}+SuperGlue~\cite{sarlin2020superglue}~{\tiny CVPR'20}  &    100\%        & 42.18     & 61.16      & 75.95   & -- \\
    DRC-Net~\cite{li2020dual}~{\tiny NeurIPS'20} &100\%    &27.01 &42.96 &58.31& --\\
    LoFTR~\cite{sun2021loftr}~{\tiny CVPR'21}&100\%     & 52.80 & 69.19  & 81.18 & 94.80 \\\hline
    MatchFormer-lite-LA   &   100\%        &48.74  &65.83   &78.81  &97.55  \\
    MatchFormer-lite-SEA    & 100\% & 48.97   &66.12   &79.07   & 97.52\\
    MatchFormer-large-LA    &  100\%        &\textbf{52.91}~\gbf{+0.11} &\textbf{69.74}~\gbf{+0.55}   &\textbf{82.00}~\gbf{+0.82}  &\textbf{97.56}~\gbf{+2.76}  \\
    \midrule 
    \multicolumn{6}{l}{\textbf{Robustness with less training data and fewer GPU resources:}} \\
    \midrule
    LoFTR\dag& 10\%    & 38.81     & 54.53      & 67.04   & 83.64 \\
    MatchFormer\dag~ &    10\%        & 42.92~\gbf{+4.11}     &  58.33~\gbf{+3.80}     & 70.34~\gbf{+3.30}   & 85.08~\gbf{+1.44}\\
    \hline
    LoFTR\dag&30\%  &  47.38    &   64.77    &  77.68  &91.94 \\
    MatchFormer\dag~ &    30\%   & 49.53~\gbf{+2.15}    &  66.74~\gbf{+1.97}    &  79.43~\gbf{+1.75}     & 94.28~\gbf{+2.34}    \\
    \hline
    LoFTR\dag&50\%    & 48.68     &  65.49     & 77.62   &92.54 \\
    MatchFormer\dag~ &    50\%       & 50.13~\gbf{+1.45}     &   66.71~\gbf{+1.22}    & 79.01~\gbf{+1.39}   &94.89~\gbf{+2.35} \\
    \hline
    LoFTR\dag&70\%    &  49.08    &  66.03     & 78.72   &93.86 \\
    MatchFormer\dag~ &    70\%        & 51.22~\gbf{+2.14}     &   67.44~\gbf{+1.41}    & 79.73~\gbf{+1.01}   &95.75~\gbf{+1.89} \\
    \hline
    LoFTR\dag&100\%  &50.85 & 67.56&79.96 &95.18 \\
    MatchFormer\dag~ &  100\%        &\textbf{53.28}~\gbf{+2.43} &\textbf{69.74}~\gbf{+2.18}   &\textbf{81.83}~\gbf{+1.87}  &\textbf{96.59}~\gbf{+1.41} \\
    \bottomrule
\end{tabular}
}
    \label{tab:outpose}
\vskip -2ex
\end{table}

\noindent\textbf{Results.}
As shown in Table~\ref{tab:outpose}, 
MatchFormer noticeably surpasses the \emph{detector-based} SuperGlue and DRC-Net, as well as the \emph{extract-to-match} LoFTR. Our MatchFormer-lite-LA model also achieves great performance. It can deliver a higher matching precision (P) with $97.55\% $, despite being much lighter.
Note that MatchFormer-large-SEA using the partially optimized SEA will raise an out-of-memory issue. Here, we recommend to use the memory-efficient LA in the high-resolution outdoor scenes. Our MatchFormer-large-LA model achieves consistent state-of-the-art performances on both metrics of AUC and P. 

\noindent\textbf{Robustness and Resource-Efficiency.}
It is reasonable to evaluate the robustness of the model when only less training data and fewer training resources are available in practical applications. Therefore, we further train MatchFormer-large-LA and LoFTR (marked with $\dag$ in Table~\ref{tab:outpose}) using different percentages of datasets and on fewer resources with 8 GPUs. First, compared to LoFTR$\dag$, MatchFormer$\dag$ obtains consistent improvements on different constrained data scales, \ie, the first $\{10,30,50,70,100\}$ percentages of the original dataset. It proves that MatchFormer has more promise in data-hungry real-world applications.
Second, training with the same $100\%$ data on different GPU resources, LoFTR$\dag$ has $(-1.95\%, -1.63\%, -1.22\%)$ performance drops at three AUC thresholds of $(5\degree, 10\degree, 20\degree)$ when using 8 GPUs instead of 64 GPUs. In contrast, MatchFormer maintains the stable and surprising accuracy, which shows that our method is more resource-friendly and easier to reproduce.

\subsection{Image Matching}
\vskip-1ex
\noindent\textbf{Metrics.} 
On the standard image matching task of HPatches sequences based on sequences with illumination or viewpoint change, we evaluate MatchFormer by detecting correspondences between pairs of input images. 
Following the experimental setup of Patch2Pix~\cite{zhou2021patch2pix},
we report the mean matching accuracy~(MMA) at thresholds from $[1,10]$ pixels, and the number of matches and features.
\begin{figure*}[t]
\begin{minipage}{0.65\textwidth}
  \includegraphics[width=1\textwidth]{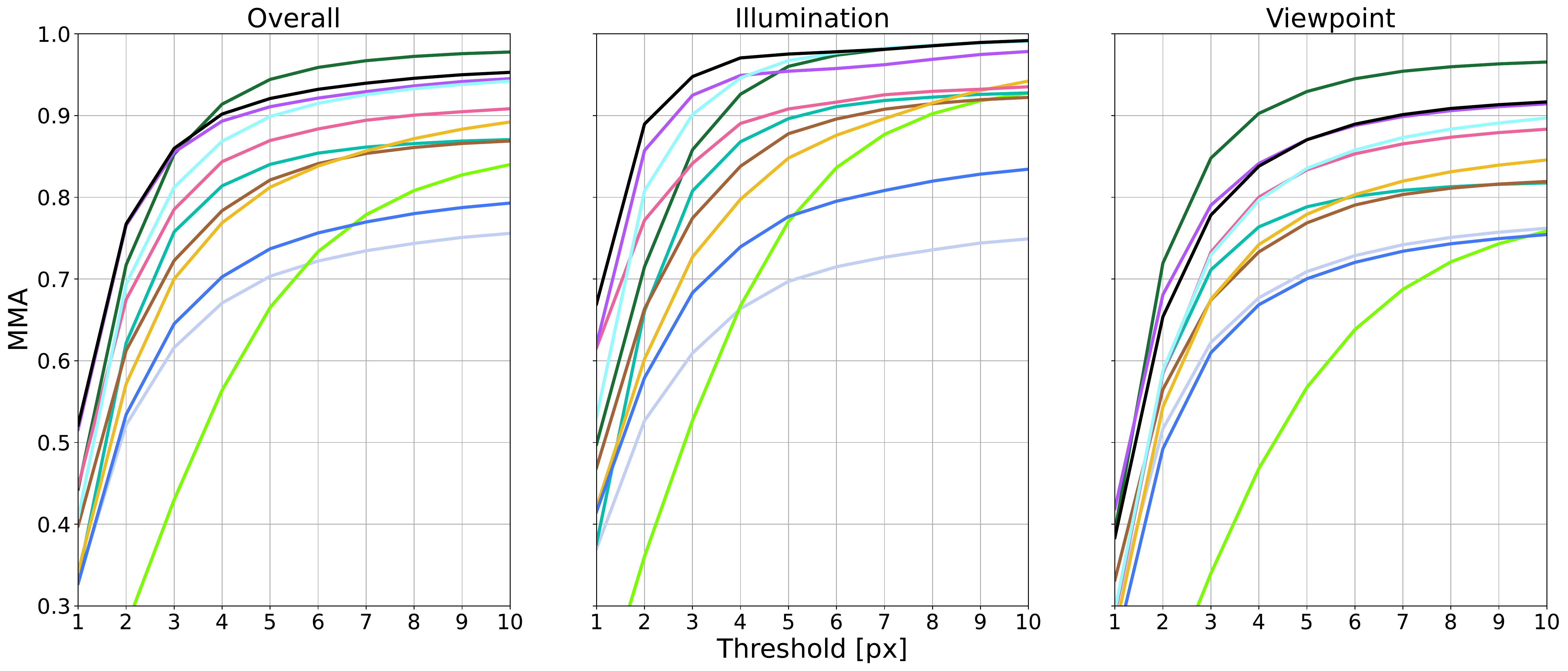}
  \vskip -2ex
\end{minipage}
\begin{minipage}{0.34\textwidth}
    \resizebox{0.95\columnwidth}{!}{
\renewcommand{\arraystretch}{1.3}
\begin{tabular}{@{}l@{}c }
    \toprule
	Methods  & \#Features / Matches\\
	\midrule
	\cstick{cdark} MatchFormer & {4.8K} / \textbf{4.8K} \\
	\cstick{csky} Patch2Pix~\cite{zhou2021patch2pix}~\tiny ICCV'21 & 1.2K / 1.2K  \\ 
	\cstick{cpurple} LoFTR~\cite{sun2021loftr}~\tiny CVPR'21 & 4.7K / 4.7K   \\
	\cstick{cgreen} SP~\cite{detone2018superpoint}+SuperGlue~\cite{sarlin2020superglue}~\tiny CVPR'20 & 0.5K / 0.9K  \\
    \cstick{cbrown} ASLFeat~\cite{luo2020aslfeat}+NN~\tiny CVPR'20 &  4.0K / 2.0K \\
    \cstick{cyellow} SP~\cite{detone2018superpoint}+ CAPS~\cite{wang2020learning}+NN~\tiny ECCV'20 &  2.0K / 1.1K \\
    \cstick{crose} SparseNCNet~\cite{rocco2020efficient}~\tiny ECCV'20 & 2.0K / 2.0K \\
    \cstick{cgrass} D2Net~\cite{dusmanu2019d2}+NN~\tiny CVPR'19 & \textbf{6.0K} / 2.5K  \\
    \cstick{ccyan} R2D2~\cite{revaud2019r2d2}+NN~\tiny NeurIPS'19 &  5.0K / 1.6K \\
 	\cstick{cpurblue} HAN~\cite{mishkin2018repeatability}+$\text{HN}_{\text{++}}$~\cite{mishchuk2017working}+NN~\tiny ECCV'18 &3.9K / 2.0K\\
    \cstick{cblue} SP~\cite{detone2018superpoint}+NN~\tiny CVPRW'18 & 2.0K / 1.1K \\

\bottomrule	
\end{tabular}
}
    \vskip -2ex
\end{minipage}  
\vskip -3ex
\caption{\small \textbf{Image matching on HPatches.} The mean matching accuracy~(MMA) at thresholds from $[1,10]$ pixels, and the number of matches and features are reported.
}
\label{fig:hpatchmatch}
\end{figure*}

\noindent\textbf{Results.} 
Fig.~\ref{fig:hpatchmatch} illustrates the results for the experiments with illumination and viewpoint changes, along with the MMA. Under varying illumination conditions, our method provides the best performance. On overall (the threshold ${\leq}3$ pixels), Matchformer performs optimally at precision levels. While other methods can only account for lighting changes or viewing angles changes, MatchFormer is reasonably compatible and maintains its functionality when the viewpoint changes. Thanks to the match-aware encoder, a larger number of features and matches, both $4.8$K, are obtained. The results reveal the effectiveness of our \emph{extract-and-match} strategy for image matching.

\subsection{Homography Estimation}
\vskip-1ex

\begin{table}[t]
    \centering
    \vskip -2ex
    \caption{\textbf{Homography estimation on HPatches.}  $\dag$ represents training on different percentages of datasets, which requires 8 GPUs for training.
    }
    \resizebox{0.99\columnwidth}{!} {
\renewcommand{\arraystretch}{0.95}
\setlength\tabcolsep{6pt}
\begin{tabular}{l | c | c c c | c}
    \toprule
	\multirow{2}{*}{Method} & {Data} & Overall & Illumination & Viewpoint  & \multirow{2}{*}{\#Matches}   \\
	 & percent & \multicolumn{3}{c|}{Accuracy ($\%,\epsilon< 1/3/5$ px)} &  \\
	\midrule\midrule
    SP~\cite{detone2018superpoint}~{\tiny CVPRW'18} & {100\%}    & 0.46/0.78/0.85 & 0.57/0.92/0.97 & 0.35/0.65/0.74   & 1.1K  \\
	D2Net~\cite{dusmanu2019d2}~{\tiny CVPR'19} & {100\%}  & 0.38/0.71/0.82 & 0.66/0.95/\textbf{0.98} & 0.12/0.49/0.67   & 2.5K  \\
	R2D2~\cite{revaud2019r2d2}~{\tiny NeurIPS'19} & {100\%} & 0.47/0.77/0.82 & 0.63/0.93/\textbf{0.98} & 0.32/0.64/0.70   & 1.6K  \\
	ASLFeat~\cite{luo2020aslfeat}~{\tiny CVPR'20}& {100\%} & 0.48/0.81/0.88 & 0.62/0.94/\textbf{0.98} & 0.34/0.69/0.78    & 2.0K  \\
	ASLFeat~\cite{luo2020aslfeat}~{\tiny CVPR'20} + ClusterGNN~\cite{shi2022clustergnn}& {100\%} & 0.51/0.83/0.89 & 0.61/\textbf{0.95}/\textbf{0.98} & 0.42/0.72/\textbf{0.82}    & -  \\
	SP~\cite{detone2018superpoint} + SuperGlue~\cite{sarlin2020superglue}~{\tiny CVPR'20}&{100\%} &  0.51/0.82/0.89 & 0.60/0.92/\textbf{0.98} &0.42/0.71/0.81  & 0.5K  \\
    SP~\cite{detone2018superpoint} + CAPS~\cite{wang2020learning}~{\tiny ECCV'20}& {100\%} & 0.49/0.79/0.86 & 0.62/0.93/\textbf{0.98}  & 0.36/0.65/0.75   & 1.1K  \\
    SP~\cite{detone2018superpoint} + ClusterGNN~\cite{shi2022clustergnn}~{\tiny CVPR'22}& {100\%} & 0.52/\textbf{0.84}/\textbf{0.90} & 0.61/0.93/\textbf{0.98}  & \textbf{0.44}/\textbf{0.74}/0.81   & -  \\
    SIFT + CAPS~\cite{wang2020learning}~{\tiny ECCV'20} & {100\%} & 0.36/0.77/0.85 & 0.48/0.89/0.95 & 0.26/0.65/0.76  & 1.5K  \\
    SparseNCNet~\cite{rocco2020efficient}~{\tiny ECCV'20} &{100\%} & 0.36/0.65/0.76 & 0.62/0.92/0.97 & 0.13/0.40/0.58  & 2.0K \\
    Patch2Pix~\cite{zhou2021patch2pix}~{\tiny CVPR'21} &{100\%} & 0.50/0.79/0.87  & 0.71/\textbf{0.95}/\textbf{0.98} & 0.30/0.64/0.76 & 1.3K \\
    LoFTR~\cite{sun2021loftr}~{\tiny CVPR'21} &{100\%}  & \textbf{0.55}/0.81/0.86  & 0.74/\textbf{0.95}/\textbf{0.98} & 0.38/0.69/0.76  & 4.7K \\
    MatchFormer&{100\%}  & \textbf{0.55}/0.81/0.87  & \textbf{0.75}/\textbf{0.95}/\textbf{0.98} & 0.37/0.68/0.78  & \textbf{4.8K}  \\
    \midrule 
    \multicolumn{5}{l}{\textbf{Robustness with less training data and fewer GPU resources:}} \\
    \midrule
    LoFTR\dag&10\%&0.50/0.78/0.84 &0.74/0.95/\textbf{0.98}&0.28/0.63/0.71&3.6K\\
    MatchFormer\dag&10\% & 0.50/0.78/0.84  & 0.72/0.93/0.97 & 0.30/0.64/0.71  & 4.0K\\ \hline
    LoFTR\dag&30\%&0.52/0.80/0.86 &0.74/0.96/\textbf{0.98}&0.32/\textbf{0.66}/0.74&4.1K\\
    MatchFormer\dag&30\% &\textbf{0.57}/\textbf{0.81}/0.86 &\textbf{0.78}/\textbf{0.97}/\textbf{0.98} &\textbf{0.36}/\textbf{0.66}/0.74& 4.4K\\ \hline
    LoFTR\dag&50\%&0.52/0.79/0.85&0.73/0.95/\textbf{0.98} &0.32/0.65/0.73&4.1K \\
    MatchFormer\dag&50\% &0.54/0.78/0.85 &0.75/0.95/\textbf{0.98} &0.35/0.62/0.74&\textbf{4.5K}\\ \hline
    LoFTR\dag&70\%&0.52/0.79/0.85 &0.74/0.94/\textbf{0.98}&0.31/0.64/0.73&4.1K\\
    MatchFormer\dag&70\% &0.55/0.79/0.86 &0.76/0.94/\textbf{0.98} &0.35/0.64/{0.75} &\textbf{4.5K}\\ \hline
    LoFTR\dag &{100\%} &0.52/0.79/0.86 &0.74/0.93/\textbf{0.98} & 0.32/0.65/0.74 & 4.2K\\
    MatchFormer\dag&{100\%}  & 0.54/0.79/\textbf{0.87}  & {0.74}/{0.95}/\textbf{0.98} & \textbf{0.36}/\textbf{0.66}/\textbf{0.77}  & \textbf{4.5K}  \\

	\bottomrule	
\end{tabular}
}
    \label{tab:exp_homography}
    \vskip -2ex
\end{table}

\noindent\textbf{Metrics.}\label{homography}
To evaluate how the matches contribute to the accuracy of the geometric relations estimation, we assess MatchFormer in the homography estimation on HPatches benchmark~\cite{balntas2017hpatches}.
The proportion of accurately predicted homographies with an average corner error distance less than $1/3/5$ pixels is reported.

\noindent\textbf{Results.}
As shown in Table~\ref{tab:exp_homography}, the large-LA MatchFormer achieves excellent performance on the HPatches benchmark in homography estimation. It reaches the best level in the face of illumination variations, delivering the accuracy of $(0.75,0.95,0.98)$ at $1/3/5$ pixel errors. Additionally, MatchFormer obtains highest number of matches with $4.8$K. 
To evaluate the robustness and resource-efficiency, we also execute experiments with varying dataset percentages in Table~\ref{tab:exp_homography}. Compared to LoFTR$\dag$, MatchFormer$\dag$ performs significantly better in homography experiments, and is relatively unaffected by the limited training data.  MatchFormer trained with $30\%$ data has a better performance in illumination variations. One reason is that the accuracy of the geometry relation estimation is related to accurate matches, as well as the distribution and number of matches~\cite{zhou2021patch2pix}.
Training with fewer GPUs on $100\%$ data, while LoFTR$\dag$ has noticeable performance drops, MatchFormer$\dag$ maintains stable performance and requires fewer training resources for success. These experiments sufficiently prove that our new \emph{extract-and-match} pipeline has higher robustness than the \emph{extract-to-match} one used in previous methods.

\subsection{Visual Localization on InLoc}
\vskip-1ex

\noindent\textbf{Metrics.}
A robust local feature matching method ensures accurate visual localization. 
To evaluate our local feature matching method MatchFormer, we test it on the InLoc~\cite{taira2018inloc} benchmark for visual localization. 
Referring to SuperGlue~\cite{sarlin2020superglue}, we utilize MatchFormer as the feature matching step to complete the visual localization task along the localization pipeline HLoc~\cite{sarlin2019coarse}. 

\begin{table}[t]
    \centering
    \caption{\textbf{Visual localization on InLoc.} We report the percentage of correctly localized queries under specific error thresholds, following the HLoc~\cite{sarlin2019coarse} pipeline. 
    }
    \resizebox{0.8\columnwidth}{!} {
\setlength\tabcolsep{8.0pt}
\renewcommand{\arraystretch}{0.8}
\begin{tabular}{l c c}
    \toprule
    \multirow{2}{*}{Method}& \multicolumn{2}{c}{Localized Queries (\%, 0.25$m$/0.5$m$/1.0$m$, 10$^\circ$)} \\
    & DUC1 & DUC2 \\
    \midrule    \midrule
    SP~\cite{detone2018superpoint} + NN~{\tiny CVPRW'18}    &  40.4 / 58.1 / 69.7  & 42.0 / 58.8 / 69.5   \\
    D2Net~\cite{dusmanu2019d2} + NN~{\tiny CVPR'19}   & 38.4 / 56.1 / 71.2  &  37.4 / 55.0 / 64.9\\ 
    R2D2~\cite{revaud2019r2d2} + NN~{\tiny NeurIPS'19}   & 36.4 / 57.6 / 74.2 & 45.0 / 60.3 / 67.9 \\    
    SP~\cite{detone2018superpoint} + SuperGlue~\cite{sarlin2020superglue}~{\tiny CVPR'20} & {49.0} / 68.7 / 80.8 & 53.4 / \textbf{77.1} / 82.4 \\ 
    SP~\cite{detone2018superpoint} + CAPS~\cite{wang2020learning} + NN~{\tiny ECCV'20}   & 40.9 / 60.6 / 72.7 &  43.5 / 58.8 / 68.7 \\ 
    SP~\cite{detone2018superpoint} + ClusterGNN~\cite{shi2022clustergnn}~{\tiny CVPR'22} & 47.5 / 69.7 / 79.8 & 53.4 / \textbf{77.1} / 84.7 \\    
    ASLFeat~\cite{luo2020aslfeat} + SuperGlue~\cite{sarlin2020superglue}~{\tiny CVPR'20} & 51.5 / 66.7 / 75.8 & 53.4 / 76.3 / 84.0 \\
    ASLFeat~\cite{luo2020aslfeat} + ClusterGNN~\cite{shi2022clustergnn}~{\tiny CVPR'22} & \textbf{52.5} / 68.7 / 76.8 & 55.0 / 76.0 / 82.4 \\
    \midrule
    SIFT + CAPS~\cite{wang2020learning} + NN~{\tiny ECCV'20} 	 & 38.4 / 56.6 / 70.7 & 35.1 / 48.9 / 58.8 \\
    SparseNCNet~\cite{rocco2020efficient}~{\tiny ECCV'20}	& 41.9 / 62.1 / 72.7 & 35.1 / 48.1 / 55.0 \\
    Patch2Pix~\cite{zhou2021patch2pix}~{\tiny CVPR'21}& 44.4 / 66.7 / 78.3  &  49.6 / 64.9 / 72.5     \\    
    LoFTR-OT~\cite{sun2021loftr}~{\tiny CVPR'21} & 47.5 / 72.2 / 84.8  &  54.2 / {74.8} / \textbf{85.5}     \\ 
    MatchFormer    			         & 46.5 / \textbf{73.2} / \textbf{85.9}  &  \textbf{55.7} / 71.8 / 81.7     \\ 
    \bottomrule	
\end{tabular}}
    \label{tab:inloc}
\vskip -4ex
\end{table}

\noindent\textbf{Results.}
As shown in Table~\ref{tab:inloc}, on the InLoc benchmark for visual localization, MatchFormer reaches a level comparable to the current state of art methods SuperGlue and LoFTR. Interleaving attention in the MatchFormer backbone enables robust local feature matching in indoor scenes with large low-texture areas and repetitive structures. 
\vskip-3ex

\subsection{MatchFormer Structural Study}
\label{sec:ablation_study}
\vskip-1ex
Performing the \emph{extract-and-match} strategy in a pure transformer, the layout between self- and cross-attention co-existing inside each stage of MatchFormer is a critical point to achieve efficient and robust feature matching. The structural study is conducted to explore the sweet spot to arrange attention modules.

\noindent\textbf{Ablation Study of Interleaving.}
To verify the rationality of the model design, models in Table~\ref{tab:ablation} are ablated according to different backbone structures, attention arrangements and patch embedding modules. Models are trained with $10\%$ data of ScanNet. Such a setting is one for efficiency and another is that the robustness between models is validated with less training data. 
By comparing \circled{1} and \circled{2}, we establish that the transformer with self-attention significantly improves the matching precision (P, ${+}5.2\%$) compared to utilizing the convolutional extractor, which shows the long-range dependency can robustify the local feature matching. 
While the structure in \circled{2} contains only self-attention in between, the structure in \circled{3} with cross-attention can bring a ${+}3.1\%$ performance gain, which demonstrates the benefits of leaning feature similarity inside a transformer. The sequential structures (\circled{4}\circled{5}) apply pure self-attention in the early stages and pure cross-attention in the later stages, while our interleaving structures (\circled{6}\circled{7}) apply mix self-/cross-attention in each stage. Our structures improve the overall performance, which adaptively inserts self-/cross-attention in multi-scale stages, and it is in line with our statement about the \emph{extract-and-match} strategy in transformers.
The comparison between \circled{4} and \circled{5} indicates that the proposed PosPE is capable of completing the fixed position encoding and it comes with a ${+}0.8\%$ gain. 
Our PosPE in \circled{7} can enhance the accuracy by ${+}0.8\%$ compared with standard PE~(StdPE) in \circled{6}, demonstrating that PosPE is more robust. Our interleaving model in \circled{7} surpasses LoFTR by a large margin~(${+}4.1\%$~@~P), indicating that MatchFormer is more robust, not only in low-texture indoor scenes, but also with less training data. 

\begin{figure}[t]
 \begin{minipage}[t]{0.55\textwidth}
  \centering
    \makeatletter\def\@captype{table}\makeatother
       \caption{\small \textbf{Ablation study} 
       with different structures, attention arrangements and PEs.
       }
       \resizebox{\columnwidth}{!}{
\setlength\tabcolsep{1pt}
\renewcommand\arraystretch{1.05}
\begin{tabular}{@{}lccccp{1.2cm}p{1.2cm}p{1.2cm}c}
    \toprule
    \multicolumn{1}{c}{\multirow{2}{*}{Method}}      &                                    \multicolumn{1}{c}{\multirow{2}{*}{Self}}     & 
    \multicolumn{1}{c}{\multirow{2}{*}{Cross}}     & 
    \multicolumn{1}{c}{\multirow{2}{*}{PosPE}}     & 
    \multicolumn{1}{c}{\multirow{2}{*}{StdPE}}     & 
    \multicolumn{3}{c}{Pose estimation AUC (\%)} &  \multicolumn{1}{c}{\multirow{2}{*}{P}} \\
    \cline{6-8}
    & & & & & @5\degree & @10\degree & @20\degree           \\
    \hline \hline
    \multicolumn{5}{l}{LoFTR~\cite{sun2021loftr}~{\tiny CVPR'21}} &   15.47& 31.72& 48.63 &82.6\\ \hline
    \circled{1} Convolution & &    &  &   &7.36& 18.17& 32.21& 76.1    \\
    \circled{2} Self-only & {\checkmark} &   &   &    & 9.48& 22.68& 38.10& 81.3    \\
    \circled{3} Cross-only & & {\checkmark}         &  & &        13.88& 29.98& 46.89& 84.4   \\ 
    \circled{4} Sequential &{\checkmark} &{\checkmark} &      &&  14.75& 31.03& 48.27& 85.0  \\ 
    \circled{5} Sequential &{\checkmark} &{\checkmark} & {\checkmark}  &&17.32 & 34.85 &52.71 &85.8 \\ 
    \circled{6} Interleaving &{\checkmark} &{\checkmark}  &  & {\checkmark} &      16.53     & 34.63 &52.31 & 85.9   \\ 
    \circled{7} Interleaving &{\checkmark} &{\checkmark}  & {\checkmark} & &\textbf{18.01}     &\textbf{35.87}       &\textbf{53.46}  &\textbf{86.7}    \\ 
    \bottomrule
\end{tabular}
}

       \label{tab:ablation}
 \end{minipage}
  \begin{minipage}[t]{0.45\textwidth}
   \centering
    \makeatletter\def\@captype{table}\makeatother
        \caption{\textbf{Efficiency analysis.} Runtime in $ms$,
        GFLOPs~@~$640{\times}480$.
        }
        \resizebox{\columnwidth}{!}{
\setlength\tabcolsep{2pt}
\renewcommand\arraystretch{1.65}
\begin{tabular}{l|cccc}
\toprule
    \multicolumn{1}{c|}{Method} & \#Params   & GFLOPs   &Runtime            & P \\
    \hline \hline
    LoFTR~\cite{sun2021loftr} &\textbf{11}             &307  &202 & 87.9\\
    LoFTR~\cite{sun2021loftr}+QuadTree~\cite{tang2022quadtree} &13 &   393  &234& 89.3\\
    MatchFormer-lite-LA &22            &\textbf{97}  & 140  &87.7 \\
    MatchFormer-large-LA &22          &389   &246    &87.8 \\
    
    MatchFormer-lite-SEA & 23             &140  & \textbf{118}  & 89.2\\
    MatchFormer-large-SEA &23             &414  &390  & \textbf{89.5}\\
    \bottomrule
\end{tabular}
}
        \label{tab:params}
 \end{minipage}
 \vskip -2ex
\end{figure}

\begin{figure}[t]
    \centering
    \includegraphics[width=1.0\linewidth]{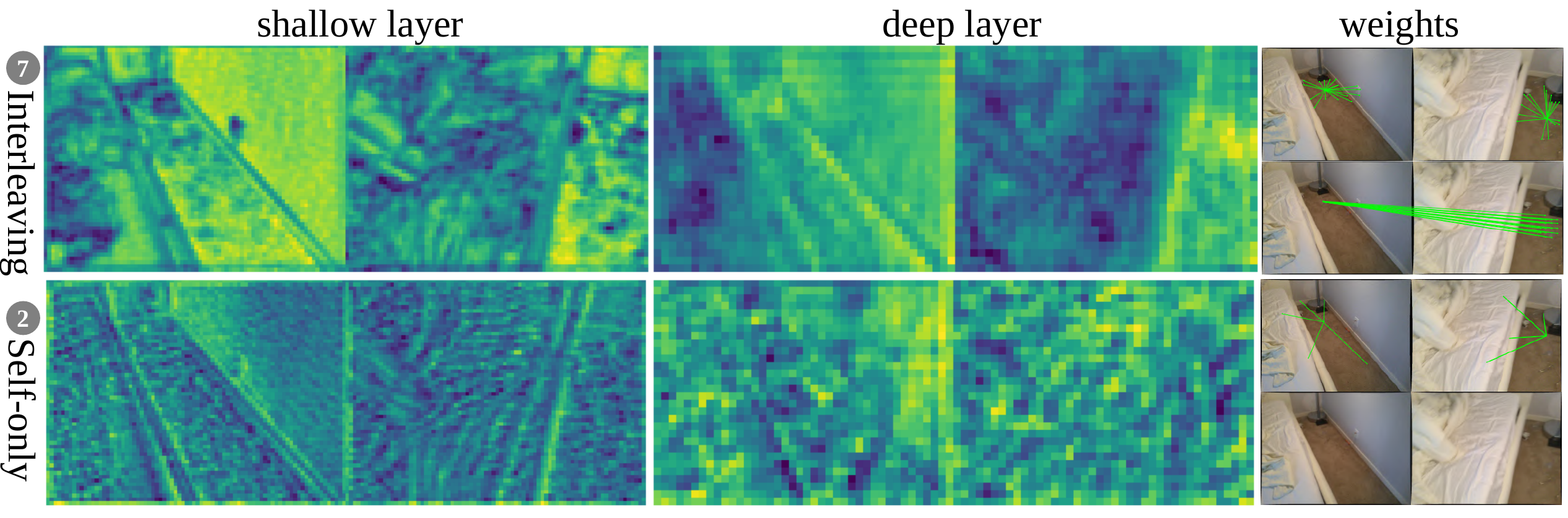}
    \vskip -3ex
    \caption{\small \textbf{Feature maps comparison} between interleaving and self-only attention in shallow and deep layers~(from the last layer of the stage-2 and stage-3, respectively).
    }
    \label{fig:attentioncompare}
    \vskip -4ex
\end{figure}

\noindent\textbf{Feature Maps Comparison.} As shown in Fig.~\ref{fig:attentioncompare}, we visualize the feature maps of the ablation experiment \circled{2} and \circled{7} of Table~\ref{tab:ablation}. 
In both shallow and deep layers, our interleaving attention structure enables MatchFormer to capture dense features and learn feature similarities, such as the paired regions highlighted in yellow. The model with only self-attention tends to extract features in each individual image and neglects the matching-aware features across images, \ie, without cross-attention weights. As a result, the self-attention model without cross-attention model will be incapable of matching local features when the image features are sparse (\ie, low-texture scenes).

\noindent\textbf{Runtime and Efficiency Analysis.}
Aside from verifying the effectiveness of arranging self- and cross-attention in an interleaving manner, MatchFormer is still supposed to be computationally efficient.
The comparisons of efficiency results including the \#Parameters ($M$), GFLOPs, and runtime ($ms$) are detailed in Table~\ref{tab:params}.
Based on a 3080Ti GPU, MatchFormer is compared against the previous transformer-based LoFTR. 
We quantify the average runtime it takes for MatchFormer to complete a single image pair on the ScanNet test set, which includes $1,500$ pairs of images in the resolution of $640 {\times} 480$. 
MatchFormer-lite-SEA is clearly much faster, speeding up the matching process by $41\%$, although a higher number of parameters is required. Additionally, we compute the GFLOPs of the two approaches to determine their computing costs and storage demands. 
The GFLOPs of MatchFormer-lite-SEA are only $45\%$ of those of LoFTR. Yet, our model achieves a $+1.3\%$ precision gain.
Thanks to interleaving self- and cross-attention in between, our
lite and large MatchFormers achieve state-of-the-art performances with respect to previous methods on various tasks.
\vskip-1ex

\section{Conclusions}
\vskip-2ex
Rethinking local feature matching from a novel \emph{extract-and-match} perspective with transformers, 
we propose the MatchFormer framework equipped
with a matching-aware encoder by interleaving self- and cross-attention for performing feature extraction and feature similarity learning synchronously.
MatchFormer circumvents involving a complex decoder as
used in the \emph{extract-to-match} methods and adopts a lightweight FPN-like decoder to fuse multi-scale features.
Experiments show that MatchFormer achieves state-of-the-art performances in indoor and outdoor pose estimation on the ScanNet and MegaDepth benchmarks, and in both homography estimation and image matching on the HPatches benchmark, as well as in visual localization on the InLoc benchmark. 

\bibliographystyle{splncs04}
\bibliography{mybibliography}

\newpage
\appendix

\section{Implementation Details}\vskip -2ex
\noindent\textbf{Transformer.}
\sloppy We design a four-stage hierarchical Transformer, using gray-scale images as input, with an input channel of $1$. Each stage contains a positional patch embedding layer and three attention layers. The channel of the feature map is gradually increased by $\{128, 192, 256, 512\}$, and the resolution is decreased by $\{1/2, 1/4, 1/8, 1/16\}$ (in the large version), or $\{1/4, 1/8, 1/16, 1/32\}$ (in the lite version). Our backbone does not contain a stem layer~\cite{he2016resnet}, and we use a large $7 \times 7$ convolution layer for the first patch embedding layer and a $3 \times 3$ convolution layer for the next three layers.

\noindent\textbf{MLP.}
Inspired by the MLP design of SegFormer~\cite{xie2021segformer}, we adopt to use the MLP layer after each attention layer in our match-aware encoder, which consists of two linear layers and a depth-wise convolution layer. 
To avoid excessive computation, we set the hidden features ratio~\cite{xie2021segformer} of all MLPs to $4$.
The MLP layers can enhance the features extracted by attention and introduce residual connections.

\noindent\textbf{Interleaving Self-/Cross-Attention.}
The \emph{extract-and-match} strategy is constructed by interleaving self- and cross-attention within our MatchFormer model. There are four stages in the match-aware encoder. As the feature map of the shallow stage (\ie, stage-1 and stage-2) emphasizes textural information, more self-attention are applied to focus on exploring the feature itself. As the feature map of the deep stage (\ie, stage-3 and stage-4) is biased toward semantic information, more cross-attention are applied to explore similarity cross images. The code of MatchFormer is reported in Algorithm~\ref{alg:code}. 

\noindent\textbf{More Structural Analysis.} To explore the effect of the attention module arrangement inside the backbone of MatchFormer, we spend large effort to analyze various self-attention and cross-attention schemes at each stage, where both modules interact in a separate or interleaved manner. To be consistent with the ablation study setting, we utilize the indoor model trained on $10\%$ of ScanNet~\cite{dai2017scannet} to conduct the experiment.  

As shown in Table~\ref{tab:structuremore}, the result in first row indicates that using only self-attention without cross-attention limits the matching capacity of transformer-based encoder.
The results of the other separate arrangements show that arranging cross-attention modules after the self-attention stage of MatchFormer can improve the performance of pose estimation, reaching $81.8\%$ in precision (P), when three stages are constructed with cross-attention modules. However, excessive usage of cross-attention will degrade the performance due to the lack of self-attention modules. Thus, we propose an attention-interleaving strategy for combining the self- and cross-attention within individual stage of backbone. In the experiments of the last four rows, the interleaving attention scheme of MatchFormer achieves the best performance ($86.7\%$ in P). The results indicate the effectiveness of our proposed interleaving arrangement and prove our observation that building a match-aware transformer-based encoder to perform the \emph{extract-and-match} strategy can benefit the feature matching.

\begin{algorithm}[t]
	\caption{\small Code of interleaving self-/cross-attention in a PyTorch-like style.}
	\label{alg:code}
	\definecolor{codeblue}{rgb}{0.25,0.5,0.5}
	\lstset{
		backgroundcolor=\color{white},
		basicstyle=\fontsize{7.2pt}{7.2pt}\ttfamily\selectfont,
		columns=fullflexible,
		breaklines=true,
		captionpos=b,
		commentstyle=\fontsize{7.2pt}{7.2pt}\color{codeblue},
		keywordstyle=\fontsize{7.2pt}{7.2pt},
	}
	\begin{lstlisting}[language=python]
	# proj: channel projection
	# DWConv: depth-wise convolution layer
	# softmax: softmax layer
	# sigmoid: sigmoid layer
	
	import torch
	import torch.nn as nn
	
	def posPE(image):
	    image = nn.Conv2D(image)
	    weight = sigmoid(DWConv(image))
	    image_enhance = image * weight
	    return image_enhance
	    
	def interleaving_attention(image_A, image_B, cross_flags):
	    seq_A, seq_B = posPE(image_A), posPE(image_B)
	    Q_A, K_A, V_A = nn.Linear(seq_A).reshape()
	    Q_B, K_B, V_B = nn.Linear(seq_B).reshape()
	    
	    for flag in cross_flags:
                if flag == True: # cross-attention
                    attn_A = Q_A @ K_B.transpose()
                    attn_A = attn_A.softmax()
                    attn_B = Q_B @ K_A.transpose()
                    attn_B = attn_B.softmax()
                    image_A = (attn_A @ V_B).transpose().reshape()
                    image_B = (attn_B @ V_A).transpose().reshape()
                        
                else: # self-attention
                    attn_A = Q_A @ K_A.transpose()
                    attn_A = attn_A.softmax()
                    attn_B = Q_B @ K_B.transpose()
                    attn_B = attn_B.softmax()
                    image_A = (attn_A @ V_A).transpose().reshape()
                    image_B = (attn_B @ V_B).transpose().reshape()
                
            return image_A, image_B
        
        # MatchFormer stages
        # stage1: cross_flags in 3 layers = [False, False, True] 
        # stage2: cross_flags in 3 layers = [False, False, True] 
        # stage3: cross_flags in 3 layers = [False, False, True] 
        # stage4: cross_flags in 3 layers = [False, False, True] 
        
        def MatchFormer(image_A, image_B):
            for _ in [stage1, stage2, stage3, stage4]:
                mage_A, image_B = interleaving_attention(image_A, image_B, cross_flags)
            return image_A, image_B
	\end{lstlisting}
\end{algorithm}
\setlength{\textfloatsep}{0.4cm}

\begin{table}[t]
    \centering
    \caption{\textbf{More Structural Analysis} of the attention arrangement in the encoder. `S' and `C' are short for a self-attention layer and a cross-attention layer, respectively.}
    \vskip -2ex
    \resizebox{\columnwidth}{!}{
\setlength\tabcolsep{10pt}
\renewcommand\arraystretch{1}
\begin{tabular}{@{}c@{}|cccccccc}
    \toprule
    &\multicolumn{4}{c}{\textbf{Structure}} & \multicolumn{3}{c}{\textbf{Pose estimation AUC}}  & \multicolumn{1}{c}{\multirow{2}{*}{P}} \\
    \cline{2-8}
    &stage1&stage2&stage3&stage4 & @5\degree & @10\degree & @20\degree \\
    \midrule \midrule
    {\multirow{6}{*}{\rotatebox[origin=c]{90}{\textit{Separate}}}} 
    &SS&SS&SS&SS &7.57     &20.57       &36.80  &75.8    \\ [-0.1cm]
    &SS&SS&SS&CC &10.77     &24.37       &42.54  &78.2    \\ [-0.1cm]
    &SS&SS&CC&CC &13.85         &30.31     &48.53     &  80.7   \\ [-0.1cm]
    &SS&CC&CC&CC &13.58         &29.57     &48.12     &  81.8   \\ [-0.1cm]
    &CC&CC&CC&CC &11.26    &26.15       &44.32  &80.9    \\  [-0.1cm]
    &SSS&SSS&CCC&CCC & 12.22 & 27.71 & 45.62 & 81.3\\ \midrule
    {\multirow{5}{*}{\rotatebox[origin=c]{90}{\textit{Interleaving}}}} 
    &SC&SC&SC&SC  &14.04     &30.57      &48.31  &81.1   \\ 
    &SSC&SSC&SSC&SSC & 12.25	&27.05	&43.78	&83.4\\
    &SCC&SCC&SCC&SCC & 14.75	&31.03	&48.27	&85.3\\
    &SSC&SSC&CCC&CCC & 12.82 & 28.48 & 46.29 & 81.0\\
    &SSC&SSC&SCC&SCC &\textbf{18.01} & \textbf{35.87} & \textbf{53.46} & \textbf{86.7}\\
    \bottomrule
\end{tabular}
}
    \label{tab:structuremore}
\vskip -2ex
\end{table}

\noindent\textbf{Coarse-to-fine Matching Module.}
The hierarchical encoder in MatchFormer extracts multi-scale features and the decoder delivers both low- and high-resolution feature pairs ($\frac{1}{r_c}$-scaled coarse features and $\frac{1}{r_f}$-scaled fine features, \textit{\wrt}, the size of input images) for coarse-to-fine matching~\cite{sun2021loftr}.

To begin with coarse matching, the $\frac{1}{r_c}$-scaled 
coarse feature pair  $\frac{H_1}{r_c}{\times}\frac{W_1}{r_c}$ and $\frac{H_2}{r_c}{\times}\frac{W_2}{r_c}$ 
is reshaped into sequences $I^c_1$ and $I^c_2$ to calculate the score $\boldsymbol{S}_{i,j}{=}\frac{1}{\tau} {\cdot} \langle I^c_1(i), I^c_2(j)\rangle$ of matrix $\boldsymbol{S}{\in}\frac{H_1W_1}{r_c}{\times}\frac{H_2W_2}{r_c}$, where $\langle\cdot,\cdot\rangle$ is the inner product, $\tau$ is the temperature coefficient, $H$ and $W$ are the image height and width. To calculate the probability of soft mutual closest neighbor matching, we use \texttt{softmax} on both dimensions of $S$ (referred to as 2D-softmax). The coarse matching probability $\boldsymbol{P}^c_{i,j}$ is calculated via 
Eq.~\eqref{scorec}. 
\begin{equation}
    \boldsymbol{P}^c_{i,j} = softmax(\boldsymbol{S}_{i,j}) \cdot softmax(\boldsymbol{S}_{j,i}).
    \label{scorec}
\end{equation}
To select coarse match predictions $\boldsymbol{M}^c$, $\boldsymbol{P}^c_{i,j}$ must be larger than the threshold $\theta$ and fulfill the mutual closest neighbor (MNN) criterion, as indicated in 
Eq.~\eqref{matchc}: 
\begin{equation}
    \boldsymbol{M}^c_{i,j} = \mathbbm{1}_{(\boldsymbol{P}^c_{i,j}{>}\theta) \;  \land \; {MNN}(\boldsymbol{P}^c_{i,j})}.
    \label{matchc}
\end{equation}

Given a matched spot $\{(i,j)|\boldsymbol{M}^c_{i,j}{=}1\}$ on coarse feature maps, its paired
windows are cropped as $(\boldsymbol{w}_{i'},\boldsymbol{w}_{j'})$ to conduct fine matching, where $(i', j')$ are back-located at the $\frac{1}{r_f}$-scaled fine feature maps. The fine match probability $\boldsymbol{P}^f_{i,j}$ of the center vector $\vec{c}_i$ of $\boldsymbol{w}_i$ related to the entire $\boldsymbol{w}_j$ can be calculated by \texttt{softmax}. 
Solving the expectation of $\boldsymbol{P}^f_{i,j} = softmax(<\vec{c}_i,\boldsymbol{w}_j>)$ can determine the fine matching $\boldsymbol{M}^f_{i,j}$ on $\boldsymbol{w}_j$, then we map it to the original resolution to establish the final matching. Fine matching can be formulated as $\mathbbm{E}_{i{\rightarrow}j}(\boldsymbol{P}^f_{i,j}|\vec{c_i})$.

\begin{table}[t]
    \centering
    \caption{\textbf{Indoor pose estimation on ScanNet with less training data.} 
    The AUC of three different thresholds and the average matching precision (P) are evaluated. 
    }
    \vskip -3ex
    \resizebox{0.99\columnwidth}{!}{
\setlength\tabcolsep{2.0pt}
\renewcommand{\arraystretch}{0.8}
\begin{tabular}{lcllll}
    \toprule
    \multicolumn{1}{c}{\multirow{2}{*}{Method}}& Data & \multicolumn{3}{c}{Pose estimation AUC} & \multicolumn{1}{c}{\multirow{2}{*}{P}} \\
    \cline{3-5} & percent & \multirow{1}{*}{@5\degree} & \multirow{1}{*}{@10\degree} & \multirow{1}{*}{@20\degree} &          \\
    \midrule \midrule
    LoFTR~\cite{sun2021loftr} & 10\% & 15.47& 31.72& 48.63 &82.6   \\
    MatchFormer-large-SEA  & 10\% & 18.01~\gbf{+2.54} & 35.87~\gbf{+4.15} & 53.46~\gbf{+4.83} & {86.7}~\gbf{+4.1} \\ \midrule
    LoFTR~\cite{sun2021loftr} & 30\% & 18.20& 35.54& 52.58 &84.1 \\
    MatchFormer-large-SEA  & 30\% & 21.20~\gbf{+3.00} & 39.65~\gbf{+4.11} & 57.16~\gbf{+4.58} & 88.5~\gbf{+4.4} \\ \midrule
    LoFTR~\cite{sun2021loftr} & 50\% & 19.65 & 37.48 & 53.89 &86.3  \\
    MatchFormer-large-SEA  & 50\% & 21.10~\gbf{+1.45}&39.91~\gbf{+2.43}&57.36~\gbf{+3.47}&89.0~\gbf{+2.7} \\ \midrule
    LoFTR~\cite{sun2021loftr} & 70\% & 19.55&37.82&54.77&85.7 \\
    MatchFormer-large-SEA  & 70\% & 21.34~\gbf{+1.79} & 41.08~\gbf{+3.26} & 58.97~\gbf{+4.20} &88.8~\gbf{+3.1}  \\ \midrule
    LoFTR~\cite{sun2021loftr} & 100\% &22.06&   40.80  & 57.62 & 87.9 \\
    MatchFormer-large-SEA  & 100\% & {24.31}~\gbf{+2.25}    & {43.90}~\gbf{+3.10}  & {61.41}~\gbf{+3.79}  & {89.5}~\gbf{+1.6}  \\ \hline
    \bottomrule
\end{tabular}
}
    \label{tab:inpose_robust}
    \vskip -2ex
\end{table}

\begin{figure}[t]
\begin{center}
    \includegraphics[width= 1.0\linewidth, keepaspectratio]{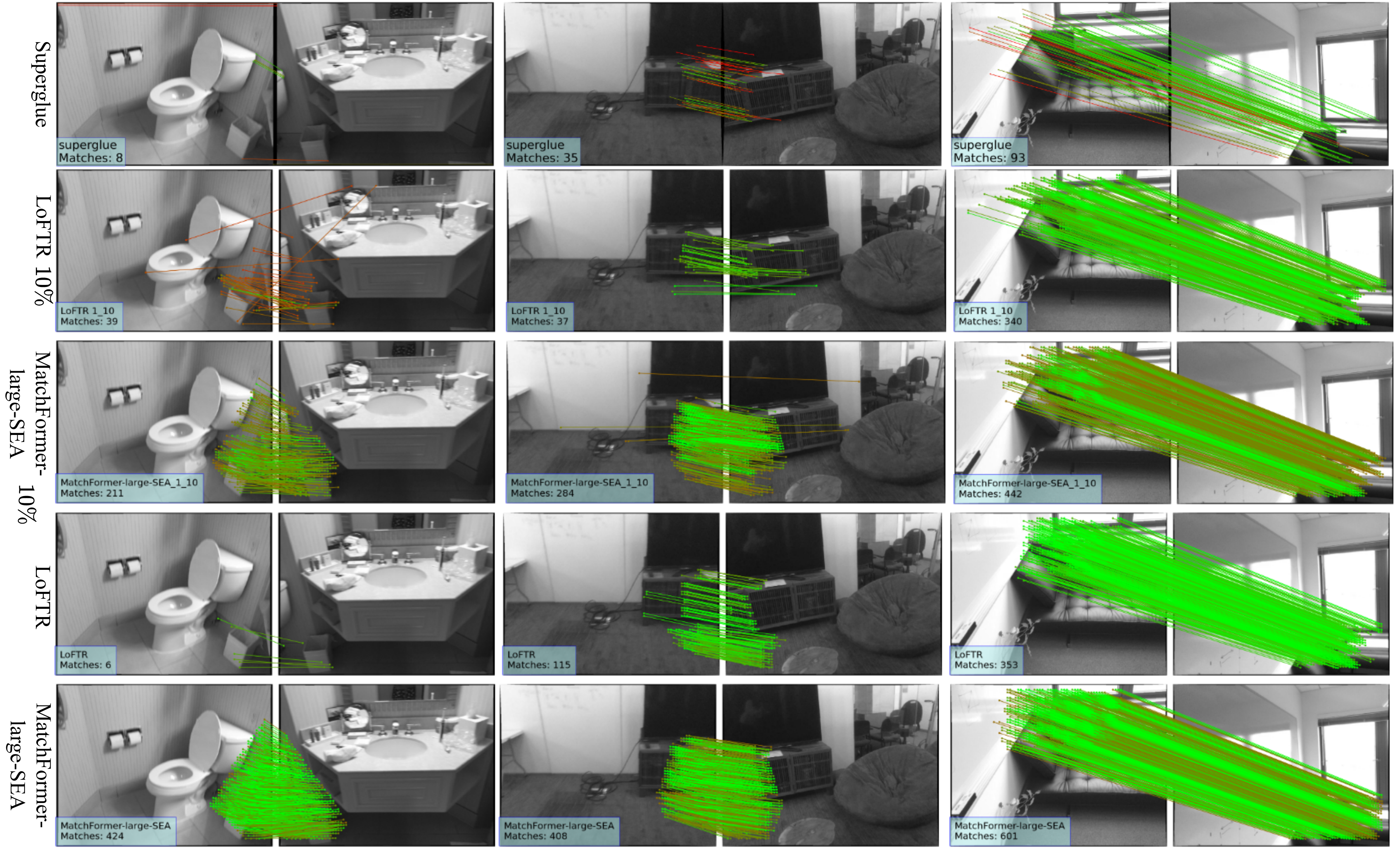}
\end{center}
   \vskip -4ex
   \caption{\small \textbf{More Qualitative Comparisons in Indoor Scene Matching} of MatchFormer, LoFTR, and SuperGlue. The color represents matching confidence, where green represents more correct matches, and red represents uncertain matches. Models ($10\%$) represent indoor models trained on $10\%$ of the ScanNet dataset~\cite{dai2017scannet}.}
\label{fig:in}
\vskip -1ex
\end{figure}
\begin{figure}[!h]
\begin{center}
    \includegraphics[width= 1.0\linewidth, keepaspectratio]{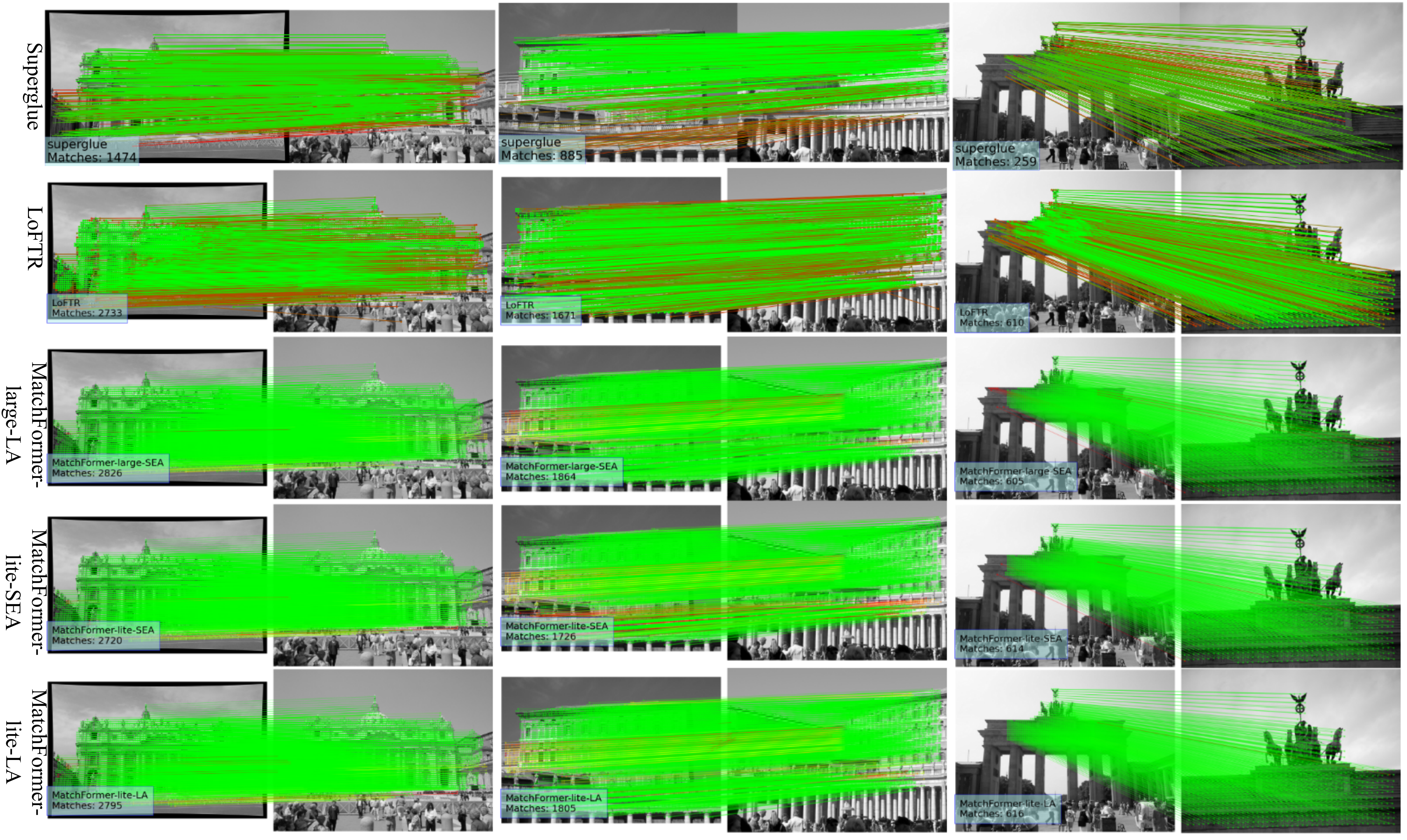}
\end{center}
    \vskip -4ex
    \caption{\small \textbf{More Qualitative Comparisons in Outdoor Scene Matching} of MatchFormer, LoFTR, and SuperGlue. The color represents matching confidence, where green represents more correct matches, and red represents uncertain matches. lite represents the model outdoor model for outputting low-resolution matching feature maps. LA represents linear attention. SEA represents spatial efficient attention.}
\label{fig:out}
\vskip -1ex
\end{figure}

\section{Indoor Pose Estimation.}\vskip -2ex

\noindent\textbf{Robustness evaluation.}
To evaluate the robustness with less training data, we further compare MatchFormer-large-LA and LoFTR in different percentages of datasets in Table~\ref{tab:inpose_robust}. The different sizes of training data are selected from the first $x{\in}\{10,30,50,70,100\}$ percentages of the original dataset.
With different sizes of training data, MatchFormer has maintained consistent performance. Hence it has tremendous potential for data-constrained real-world scenarios.

\noindent\textbf{Qualitative Comparisons.} The visualizations of indoor matching qualitative comparisons can be found in Fig.~\ref{fig:in}. From top to bottom are the matching results from SuperGlue~\cite{sarlin2020superglue}, LoFTR~\cite{sun2021loftr} with $10\%$ training data, MatchFormer-large-SEA with $10\%$ training data, LoFTR and MatchFormer-large-SEA with all training data. Due to the captured long-range dependency, MatchFormer achieves dense feature matching in such challenging indoor scenes with large viewing angle changes, such as the first and the second column in Fig.~\ref{fig:in}. In the low-texture scene of the third column, our method can provide more matches compared to SuperGlue and LoFTR.
Additionally, the performance of MatchFormer-large-SEA is significantly better than LoFTR, when they are trained on the same $10\%$ data of ScanNet, which indicates that our model is more flexible when transferred to a moderate dataset.

\begin{figure*}[t]
\begin{center}
    \includegraphics[width= 1.0\linewidth, keepaspectratio]{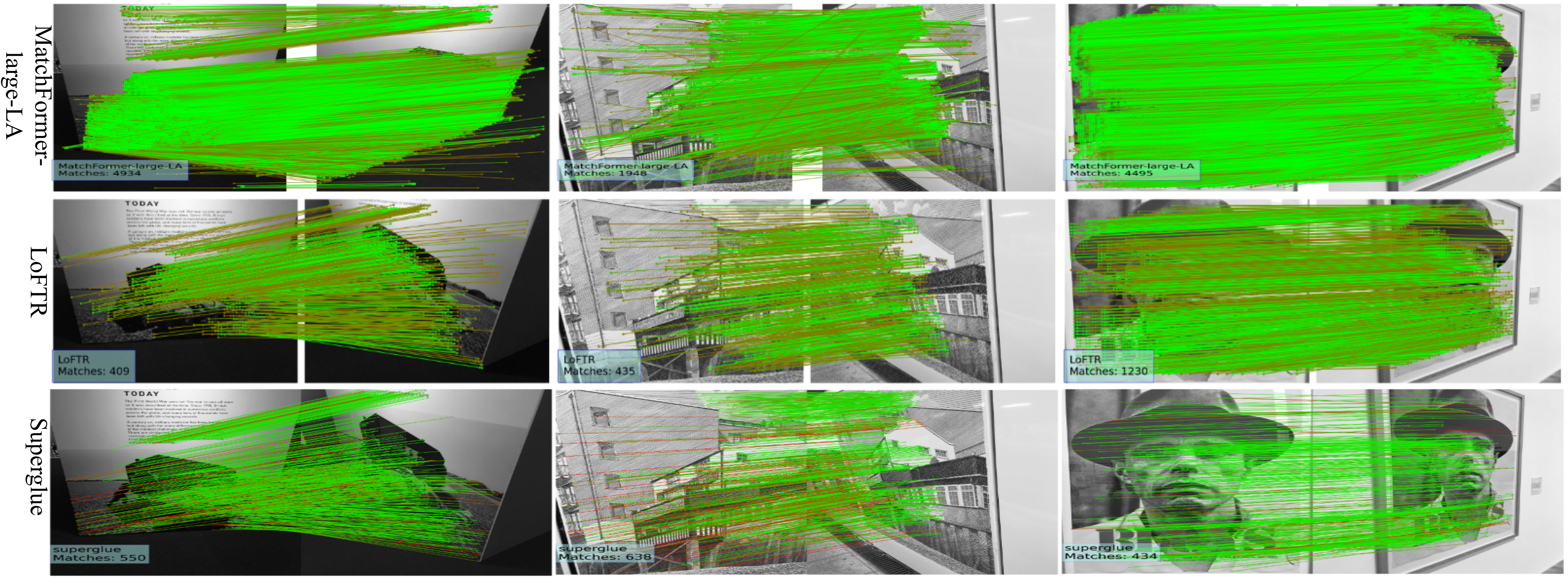}
\end{center}
    \vskip -5ex
   \caption{\small \textbf{Qualitative Comparisons on HPatches}.} 
\label{fig:hp}
\vskip -2ex
\end{figure*}
\begin{figure*}[!h]
\begin{center}
    \includegraphics[width= 1.0\linewidth, keepaspectratio]{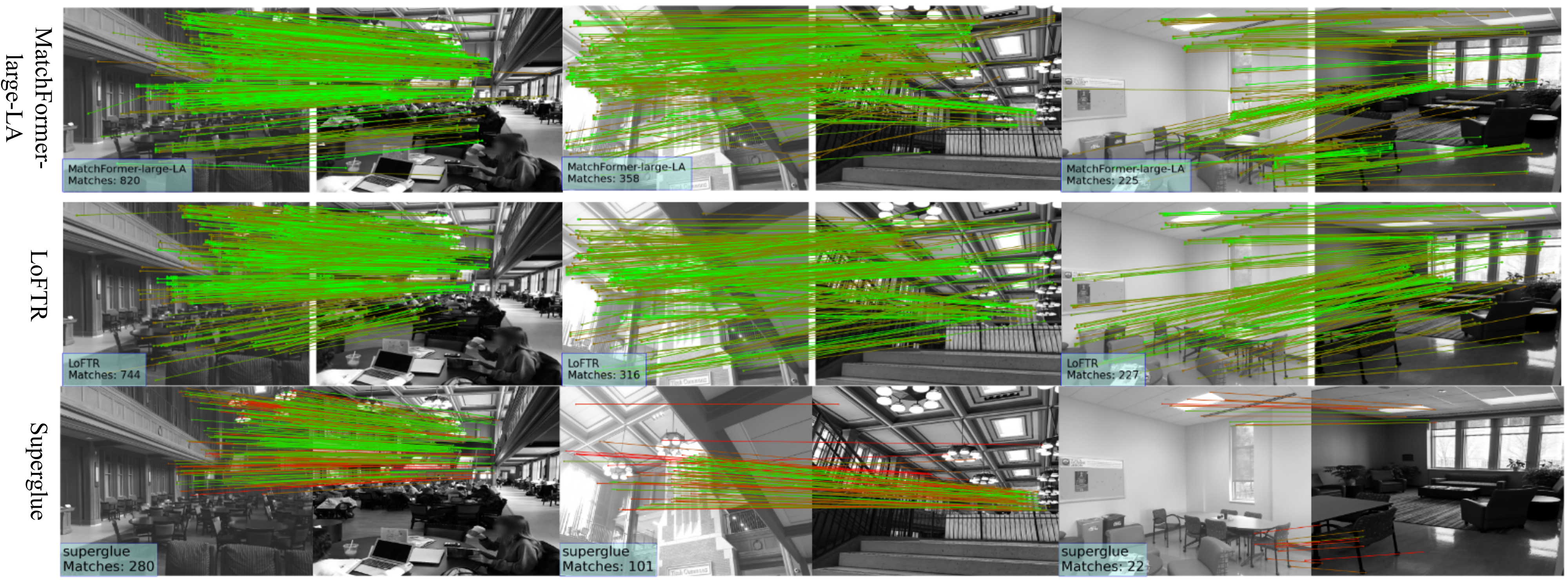}
\end{center}
    \vskip -5ex
   \caption{\small \textbf{Qualitative Comparisons on InLoc}.}
\label{fig:inloc}
\vskip -2ex
\end{figure*}

\section{Outdoor Pose Estimation}\vskip -2ex
\noindent\textbf{Qualitative Comparisons.}  As shown in Fig.~\ref{fig:out}, we visualize the qualitative comparisons of the outdoor model at MegaDepth~\cite{Li_2018_CVPR}. 
In outdoor scene matching, MatchFormer-large-LA outperforms LoFTR and SuperGlue in matching performancec.
The matching performance of MatchFormer-lite-SEA and MatchFormer-lite-LA are on par with that of LoFTR and SuperGlue. 

\section{Homography Estimation}\vskip -2ex

\noindent\textbf{Qualitative Comparisons.} To evaluate the feature matching in the benchmark for geometric relations estimation, we perform Homography Estimation on HPatches~\cite{balntas2017hpatches} with the MatchFormer-large-LA. 
In Fig.~\ref{fig:hp}, we visualize more qualitative comparison based on the matching results of MacthFormer-large-LA, LoFTR~\cite{sun2021loftr}, and SuperGlue~\cite{sarlin2020superglue}. 
MatchFormer can perform more dense and confident matching than SuperGlue. Besides, MatchFormer has further improvements by yielding more matches compared to LoFTR, such as an improvement with more than $4.5K$ matches in the first column of Fig.~\ref{fig:hp}.

\section{Image Matching}\vskip -2ex
Following the experimental setup of Patch2Pix~\cite{zhou2021patch2pix},
we choose the same $108$ HPatches sequences, including $52$ sequences with illumces with viewpoint change. Each sequence contains six images. To match the first with all others, we report the mean matching accuracy~(MMA) at thresholds from $[1,10]$ pixels, and the number of matches and features. The input size of the image is set to $1024$, the matching threshold is set to $0.2$, and RANSAC threshold as $2$ pixels.

\section{InLoc Visual Localization}\vskip -2ex
\noindent\textbf{Detailed Settings.} 
On the InLoc~\cite{taira2018inloc} benchmark, we follow Patch2pix~\cite{zhou2021patch2pix} to evaluate the same first 40 retrieval pairs. The same temporal consistency check is performed to limit the retrievals, and the RANSAC threshold is set to 48 pixels for pose estimation. We adjust the images to 1024 on the long side.

\noindent\textbf{Qualitative Comparisons.}
To evaluate the effectiveness of MatchFormer in the visual localization task, we evaluate MatchFormer-large-LA on the InLoc~\cite{taira2018inloc} benchmark. The visualizations of InLoc visual localization results can be found in Fig.~\ref{fig:inloc}. In comparison to the detector-based MatchFormer method, MatchFormer has a greater and more accurate number of matches. MatchFormer performs at a level comparable to the detector-free method LoFTR.

\section{Limitations and Future Work}\vskip -2ex
For indoor scenes and outdoor scenes, MatchFormer employs two kinds of attention, \ie, spatial efficient attention~(SEA) and linear attention~(LA), which have varying degrees of computational reductions and different abilities for feature extraction. 
They are appropriate for either indoors or outdoors. 
In our experiments, LA proved to be more suitable for outdoor scenes with dense high-resolution input. In contrast, SEA was more appropriate for indoor scenes with sparse low-resolution input. 
Exploring a uniform efficient attention module to handle both indoor and outdoor inputs with different resolutions, we leave it as the future work.
Besides, in MatchFormer, we introduce an efficient FPN-like decoder that can combine match-aware feature maps generated by interleaving attention. 
It is potential to adapt an alternative decoder to the feature fusion task, such as MLP-decoder.

\section{Acknowledgments}\vskip -2ex
This work was supported in part by the Federal Ministry of Labor and Social Affairs (BMAS) through the AccessibleMaps project under Grant 01KM151112, in part by the University of Excellence through the ``KIT Future Fields'' project, in part by the Helmholtz Association Initiative and Networking Fund on the HAICORE@KIT partition, and in part by Hangzhou SurImage Technology Company Ltd.

\end{document}